\documentclass[acmlarge,screen,nonacm]{acmart}
\usepackage{listings}

\lstdefinestyle{jsonstyle}{
  basicstyle=\footnotesize\ttfamily,
  columns=fullflexible,
  breaklines=true,
  breakindent=2em,
  showstringspaces=false,
  frame=single,
  framesep=5pt,
  xleftmargin=5pt,
  xrightmargin=5pt,
  aboveskip=6pt,
  belowskip=2pt,
  literate={–}{{-}}1 {—}{{---}}1
}

\usepackage{subcaption}

\usepackage{booktabs}   
\usepackage{siunitx} 
\usepackage{float}
\usepackage{makecell}
\usepackage{multirow}

\usepackage[most]{tcolorbox}
\newtcolorbox[auto counter]{simplejsonbox}[1][]{%
  colback=gray!5,
  colframe=gray!60!black,
  fonttitle=\bfseries\small,
  title={#1},
  boxrule=0.4pt,
  arc=1mm,
  left=4pt, right=4pt, top=2pt, bottom=2pt,
}

\lstdefinestyle{promptstyle}{
    basicstyle=\ttfamily\small,
    breaklines=true,
    breakatwhitespace=false,
    columns=fullflexible,
    frame=single,
    keepspaces=true,
    showstringspaces=false,
    tabsize=2
}

\AtBeginDocument{%
  }

\setcopyright{none}
\copyrightyear{2026}
\acmYear{2026}

\acmJournal{JOCCH}

\begin{document}

\title{From Classification to Generation: Vision--Language Models for Roman Imperial Coin Description}

\author{Pedro Lourenco}
\affiliation{%
  \institution{School of Computer Science, University of St Andrews}
  \city{St Andrews}
  \country{United Kingdom}
}
\email{pl93@st-andrews.ac.uk}

\author{Ognjen Arandjelovi\'c}
\affiliation{%
  \institution{School of Computer Science, University of St Andrews}
  \city{St Andrews}
  \country{United Kingdom}
}
\email{oa7@st-andrews.ac.uk}

\renewcommand{\shortauthors}{Lourenco and Arandjelovi\'c}

\begin{abstract}
Cataloguing a Roman Imperial coin involves reading its legends and identifying its imagery in order to match it to an entry in the standard reference catalogue. Previous work has automated individual parts of this process, such as identifying the issuer or recognizing reverse motifs, by formulating each part as a  classification task over a predefined set of labels. We instead formulate coin description as image-to-text generation and adapt an open-weight vision--language model (VLM), Qwen3-VL-8B-Instruct, to four tasks: transcribing the obverse and reverse legends, and describing the obverse and reverse imagery. To obtain training data, we convert approximately 30,000 auction records into structured image--text pairs. We separate the two sides of each coin, extract and normalize legends and descriptions, and standardize obverse descriptions against a bust-type taxonomy. We then train a separate low-rank adapter for each task on a single 20\,GB GPU, and evaluate each task with metrics that we define to capture the key components of numismatic descriptions. The adapted models outperform both the unadapted base model and GPT-4.1 on all four tasks, reducing the normalized edit distance of legend transcriptions by over 40\% relative to GPT-4.1. The improvement is smaller, but still present, on legends absent from the training data.
\end{abstract}


\maketitle

\section{Introduction}\label{sec:introduction}
\begin{figure}[t]
    \centering
    \includegraphics[width=0.32\linewidth, keepaspectratio]{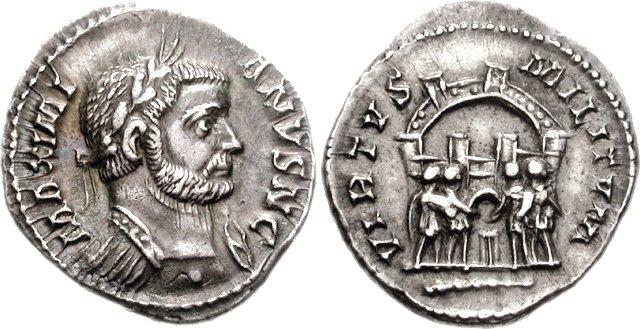}\hfill
    \includegraphics[width=0.32\linewidth, keepaspectratio]{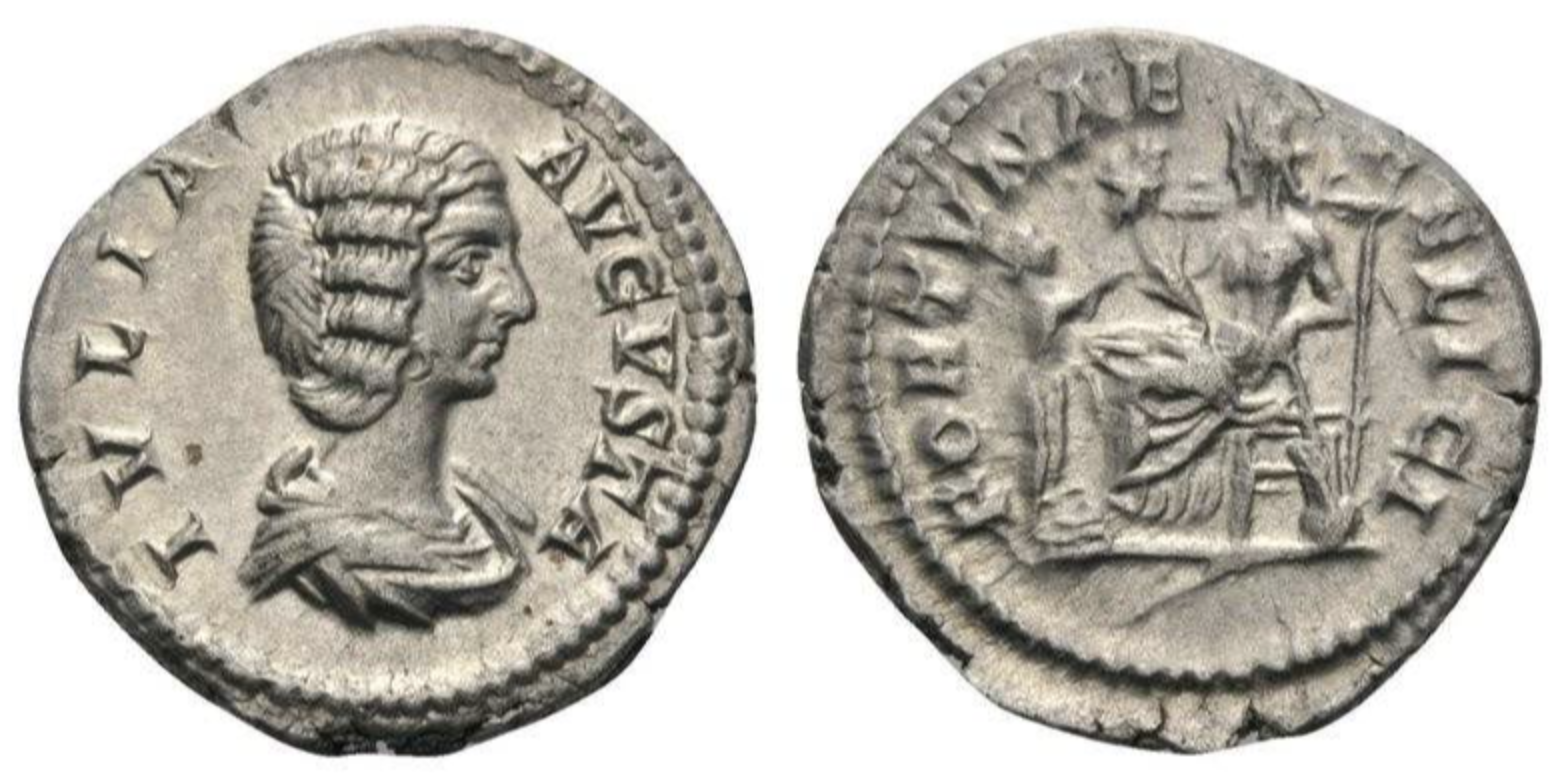}\hfill
    \includegraphics[width=0.32\linewidth, keepaspectratio]{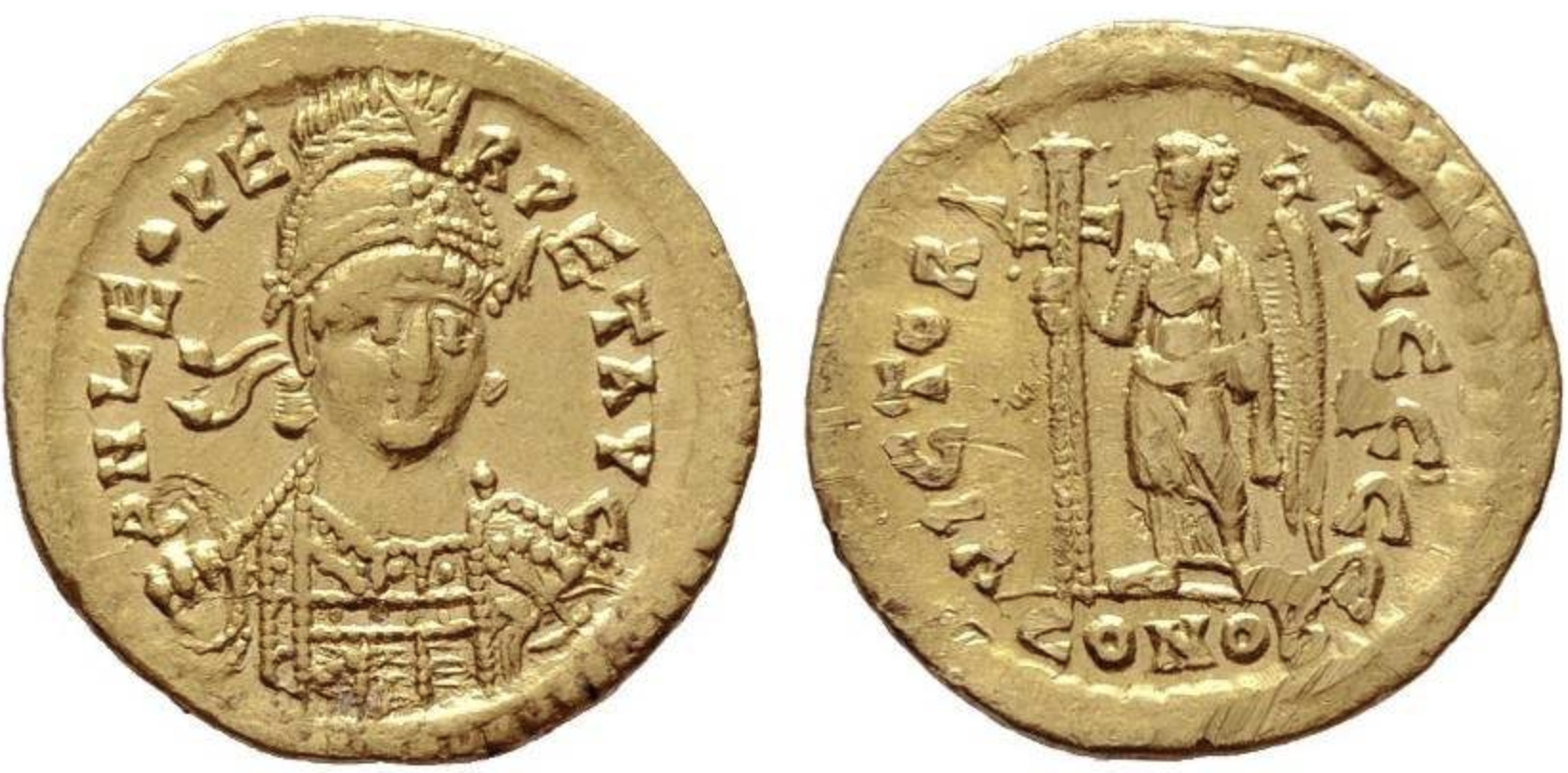}
    \caption{Sample images from the dataset, each showing both sides of a single coin.}
    \label{fig:sample-roman-coin-image}
\end{figure}

The aim of the present work is to automate the transcription of inscriptions and the description of motifs on Roman Imperial coins from their photographs. Our focus on this coinage in particular -- as opposed to, say, Ancient Greek, Celtic, or Byzantine -- is motivated by the fact that it features the greatest diversity of content, both textual and pictorial, while also being well documented in standard references such as Roman Imperial Coinage (RIC)~\citep{sutherland1984ric}. The said diversity is also usefully structured for systematic study: the obverse of a coin (colloquially: the heads) usually bears a portrait of the issuing authority (usually the emperor or the caesar) and an inscription, in numismatic jargon referred to as the legend, which commonly records the depicted person's name and titles; the reverse is highly variable, presenting pictorial motifs that range from deities and votive objects, to scenes of battles and architectural feats, accompanied by a further legend; see Figure~\ref{fig:sample-roman-coin-image}. This structure and the conventions that underlie it provide a standardized and widely used framework for description, while the variety of legends and designs makes the task sufficiently difficult to be both of practical interest and a challenge to machine learning methods that extends far beyond the recognition across a small set of predefined categories. The practical interest specifically stems from the demandingness of the task that requires specialist knowledge and considerable time investment -- automating the process would support professionals and amateurs alike, both in identification through comparison with documented catalogue types and the recording of specimens for which no matching type is available.

What makes the problem challenging has been well established by previous work~\cite{arandjelovic2010,zaharieva2007,zambanini2013,arandjelovic2020images}, so here we summarize a few key factors. First, specimens of the same type can differ substantially in appearance because of variation in die engraving, off-centre striking, wear, and corrosion~\citep{fare2017ancient}. On the other hand, types recognized as distinct by numismatic experts may differ only in a few letters or a small pictorial detail. This means that it does not suffice to recognize the portrait depicted or the broad content of the reverse scene -- a nuanced and sufficiently complete understanding of the entirety of a coin's contents is usually necessary to pinpoint its specific type, which just on the obverse includes distinguishing the identities, poses, clothing, and held objects that characterize the design.

Previous work has attempted to tackle a range of different, specific challenges noted above, including obverse issuing authority identification~\cite{arandjelovic2010,schlag2017}, legend recognition~\cite{arandjelovic2012,kavelar2014}, catalogue-type classification~\cite{kim2016,guo2023}, and reverse-motif recognition~\cite{cooper2020,anwar2021,reid2026}. Importantly, these methods generally formulate their tasks as predictions of labels from predefined label sets or as specimen identification by comparison with a gallery of reference images. While at first sight reasonable, both approaches are practically significantly limited because they rely on the availability of exhaustive reference class sets and exemplars. This is a significant constraint in a domain containing tens of thousands of coin types, many represented by few photographed specimens~\citep{arandjelovic2020images,cooper2020}. The few works that instead approach the problem through the recognition of individual semantic elements~\citep{cooper2020} offer greater flexibility but are still overly constrained by incomplete motif coverage and the limited number of relevant labels. The method we present, of generating legends and descriptions directly, allows for the expression of information beyond a fixed set of complete inscriptions or catalogue entries, making the approach relevant also to specimens whose type is absent from reference collections and facilitating matching across both textual and visual modalities. We outline the approach and summarize the key contributions next, before proceeding with an in-depth description of the present proposal.

Vision--language models (VLMs)~\cite{liu2024} allow automated coin description to be approached differently. Rather than predicting a separate label for each component, a VLM takes an image of the coin as input and generates text, so it can produce legend transcriptions and descriptions of the imagery on the coin directly. Without adaptation, however, general-purpose VLMs perform poorly on the task of automated coin description, as Section~\ref{sec:results} shows. Coin surfaces are often worn or corroded, small details such as a single letter or an object held by a figure can change the correct description, and numismatic descriptions use specialized terminology. Material of this kind is unlikely to be well represented in the data on which general-purpose VLMs are trained. 

Supervised fine-tuning (SFT), in which a pre-trained model is trained further on example inputs paired with the desired outputs, has been used to adapt general-purpose VLMs to specialized domains~\cite{li2023llavamed,kuckreja2024,semnani2025} and offers a way to address this domain gap. However, suitable training data are not readily available. Large collections of coin photographs exist~\cite{acsearch}, but they are typically accompanied by auction descriptions that are unstructured, multilingual, and inconsistently formatted, rather than by structured targets for legend transcription and imagery description. Converting these records into suitable training data therefore requires substantial preprocessing and standardization.

In this work, we construct a structured corpus of roughly 30,000 Roman Imperial coins from auction records and use it to adapt Qwen3-VL-8B-Instruct~\cite{bai2025}, an open-weight VLM, to four tasks: obverse legend transcription, reverse legend transcription, obverse description generation, and reverse description generation. We define task-specific evaluation metrics for each output type and train a separate adapter for each task using parameter-efficient adaptation~\cite{hu2021,dettmers2023}, with the model loaded in 4-bit precision on a single 20\,GB GPU. We evaluate the adapted models against both the unadapted Qwen3-VL-8B-Instruct and GPT-4.1~\cite{openai2025gpt41}, and additionally test legend transcription on inscriptions absent from the training data.

The contributions of this work are as follows.
\begin{itemize}
\item A generative formulation of automated Roman Imperial coin description, replacing closed-set classification with autoregressive text generation across four tasks: obverse and reverse legend transcription, and obverse and reverse description generation.

\item A data construction pipeline that converts noisy, multilingual auction records into a structured image--text corpus of roughly 30,000 coins. The pipeline separates the two coin faces, extracts and normalizes their legends and descriptions, translates non-English descriptions, and standardizes obverse descriptions against a bust-type taxonomy.

\item  Task-specific evaluation metrics for each of the tasks we defined, measuring how closely the generated legend transcriptions and imagery descriptions match the reference annotations.

\item An empirical evaluation showing that parameter-efficient adaptation of Qwen3-VL-8B can be performed on a single 20\,GB GPU and substantially improves performance across all four tasks. The adapted models outperform both the unadapted base model and GPT-4.1, with improvements in legend transcription persisting on legends absent from the training data.
\end{itemize}

\section{Related Work}\label{sec:related-work}
Early approaches to automatic ancient-coin classification relied on hand-engineered visual features. \citet{zaharieva2007} investigated whether methods developed for modern coin recognition could be transferred to ancient coins, including early experiments with the Scale-Invariant Feature Transform (SIFT)~\cite{lowe1999}, an approach developed further by \citet{kampel2008}. \citet{arandjelovic2010} later developed a more specialized approach for Roman Imperial coins, introducing locally-biased directional histograms to preserve the spatial relationships between local features, together with automatic coin localization and segmentation. On a 65-class dataset, where classes corresponded to the emperor depicted on the obverse, the proposed representation achieved 57.2\% classification accuracy, compared with 2.4\% for a geometry-free histogram of SIFT visual words.   

Subsequent work brought legends into the coin type classification pipeline, with \citet{arandjelovic2012} introducing one of the first methods to extract a legend directly from a photograph of an ancient coin. The extracted legend was used to find candidate coin types from online coin repositories, and images of the candidates' reverses were then matched against the reverse of the coin being classified to determine its type. \citet{zambanini2013} combined exemplar-based image matching with lexicon-based legend recognition for the classification of Roman Republican coin types, and showed that the combined approach outperformed either source of information alone. \citet{kavelar2014} then addressed legend recognition as a task in its own right. They observed that conventional optical character recognition (OCR) is poorly suited to coin legends because the lettering and background have the same colour, making standard text-background segmentation unreliable. Instead, letters were treated as visual objects: each candidate letter location was described with SIFT descriptors and scored by letter classifiers. The best-scoring letter configurations were then matched against a lexicon of known words.

The next major shift was towards deep learning, initially through convolutional neural networks (CNNs). \citet{kim2016} trained CNN classifiers to predict the RIC type of Roman Imperial coins, exploiting the fact that each RIC type belongs to exactly one emperor. One CNN takes the obverse image and outputs a probability for each emperor; a second takes the reverse image and outputs a probability for each RIC type. The two are combined by multiplying the probability of a type by that of its emperor, with combinations in which the type does not belong to the emperor set to zero, and the type with the highest product is assigned to the coin. They also analysed which image regions were most important to the classifier and found that these largely coincided with landmarks used by human experts. \citet{schlag2017} narrowed the problem further, focusing exclusively on the issuing authority, the emperor or empress, depicted on the obverse. Exploiting the fact that Roman Imperial coins almost universally carry this portrait in profile, they formulated the task as profile-face recognition and designed a CNN specifically for it. Using substantially larger and more systematically constructed datasets than earlier work, including one graded by coin condition, they showed that this specialized approach outperformed previous generic local-feature methods by roughly an order of magnitude.

\citet{guo2023} observed that classification-based methods must be retrained whenever images of a new class become available. This is impractical for Roman Imperial coinage, which has tens of thousands of catalogue entries, many with few or no available images. They therefore replaced classification with pairwise matching. Their model, a Siamese network built on vision transformers, takes images of two coins and outputs a score between 0 and 1, which the authors treat as a quasi-probability that the two coins belong to the same catalogue entry. At inference time, a query coin is compared against a reference gallery of coins with known catalogue entries, and if the best match scores highly enough, the query is assigned the matching coin's entry; otherwise, the gallery coins are returned ranked by score. New catalogue entries can thus be supported without retraining, simply by adding an example of each to the gallery.

A different line of work has focused on recognizing the motifs depicted on coin reverses. Building on earlier bag-of-visual-words motif recognition~\citep{anwar2015}, \citet{anwar2021} introduced CoinNet, which frames this as a multi-class classification problem: given an image of a Roman Republican coin’s reverse, the model assigns it to exactly one class from a fixed set of reverse motifs. CoinNet extracts features with two convolutional networks pre-trained on ImageNet, fuses them using compact bilinear pooling, and passes the result through residual blocks and a soft-attention layer. \citet{cooper2020} addressed the same task with a different formulation and a different source of supervision. Instead of assigning each reverse to a single motif class, they trained a separate CNN as a binary classifier for each of five common motifs, to predict whether that motif is present. Instead of a hand-curated dataset, they derived labels by keyword-matching the unstructured auction descriptions accompanying 100,000 coin images from ACSearch, a source also used in the present work. \citet{reid2026} extended this approach to eight motifs and replaced the per-motif CNNs with a single multi-label vision transformer trained on the same kind of automatically labelled data. The transformer outperformed CNNs retrained for the comparison, although its accuracy was within a few points of that originally reported by \citet{cooper2020}. This use of auction descriptions as automatically derived supervision is closely related to the data construction strategy adopted in the present work.

Across this literature, each method produces outputs from a set defined in advance: classifiers select from predefined labels, lexicon-based methods recognize words from a known vocabulary, and matching approaches compare a coin against a reference gallery. To our knowledge, autoregressive text generation, in which a model produces its output one token at a time, has not yet been explored for transcribing legends or describing motifs on Roman coins. In this work, we adopt this formulation for both tasks. This formulation is enabled by modern VLMs, which combine transformer-based language models~\citep{vaswani2017} with visual representations, commonly produced by vision transformers~\citep{dosovitskiy2021}, to generate text conditioned on an image~\citep{liu2024}. General-purpose VLMs, however, often perform poorly when applied directly to specialized domains, and domain-specific adaptation has been explored in biomedical imaging~\citep{li2023llavamed}, remote sensing~\citep{kuckreja2024}, and historical text recognition~\citep{semnani2025}. In each case, the authors attribute the shortcomings of general-purpose models to differences between the domain material and the data on which those models were trained, and address them by training on domain-specific data. We observe a similar gap for Roman Imperial coins and address it through supervised fine-tuning. The availability of open-weight VLMs and of parameter-efficient methods such as low-rank adaptation (LoRA)~\citep{hu2021} and its quantized variant QLoRA~\citep{dettmers2023} makes such adaptation of billion-parameter models feasible on modest hardware.

\section{Methodology}\label{sec:methodology}
\subsection{Task Formulation}
We formulate the task of describing a Roman Imperial coin as four image-to-text generation tasks. All four tasks share the same pre-trained VLM, whose weights remain frozen, and a separate set of LoRA adapters is trained for each task. The four tasks are:
\begin{itemize}
    \item \textbf{Obverse legend transcription:} given an image of the coin's obverse, generate the obverse legend.

    \item \textbf{Reverse legend transcription:} given an image of the coin's reverse, generate the reverse legend.

    \item \textbf{Obverse description generation:} given an image of the coin's obverse, generate a description of the obverse portrait and its attributes.

    \item \textbf{Reverse description generation:} given an image of the reverse side of the coin, generate a description of the reverse imagery.
\end{itemize}

\subsection{Dataset Construction}\label{sec:dataset}
\subsubsection{Source Data and Filtering}
The data used in this work were provided by the Ancient Coin Search Engine (ACSearch)~\citep{acsearch}, an online database of ancient coins offered at auction. Each record contains a high-quality photograph of a coin, accompanied by a free-text description of the coin provided by the auction house that offered it. Each photograph typically shows both sides of a coin against a uniform background, as shown in Figure~\ref{fig:sample-roman-coin-image}, while the accompanying free-text descriptions vary greatly in terms of structure and format.

Two properties of this archive mean that it required further processing before it could be used to fine-tune models for the four tasks we have described. The first concerns coverage: the archive spans Greek, Celtic, Byzantine, and Roman coinage, whereas this work focuses on Roman Imperial coins only. Records were therefore filtered by regular-expression matching and kept only if the auction description cited the \emph{Roman Imperial Coinage} catalogue (e.g.\ ``RIC 123'', ``R.I.C.\ 123'', ``RIC IV 123''). The second concerns the form of the auction descriptions: they contain the information we need, but as unstructured, multilingual, and inconsistently formatted text rather than as separate fields that can serve directly as training targets for each of the four tasks. Converting the free-text descriptions into these fields is described in Section~\ref{sec:structured-label-extraction}.

\subsubsection{Structured Label Extraction}\label{sec:structured-label-extraction}
Each description was converted into a structured representation in two stages: a generative extraction stage, followed by a deterministic validation and enrichment stage.

The extraction stage used OpenAI's \texttt{gpt-4.1-mini} model~\citep{openai2025gpt41mini}, accessed through the Responses API with a temperature of 0 and structured outputs~\citep{openai2024structured}. Each request consisted of two messages. The system message contained the prompt shown in Appendix~\ref{appendix:structured-features-extraction-prompt}, which defines the task, and sets out rules for assigning text to the obverse and reverse, copying legends exactly as written, and translating descriptions into English. The user message contained the full auction description, together with a single photograph showing both sides of the coin. The structured-output schema constrained the response to a JSON object with exactly four string fields: obverse legend, obverse description, reverse legend, and reverse description.

After the extraction stage, we applied deterministic steps to the model's output to increase its reliability. First, a legend was accepted only if it appeared as a substring of the source auction description, after both were converted to upper case and every run of white-spaces was collapsed to a single space. Otherwise, it was set to \texttt{unknown}, which discarded any legend that the model did not copy verbatim from the description. Second, accepted legends are normalized into a canonical form by removing all characters other than letters, and mapping \textsc{u} to \textsc{v}, following the convention that Latin inscriptions use a single letter for both. 

The enrichment stage added a simplified catalogue identifier to each record. We first curated a list of Emperors and Augustae, together with their common aliases. The aliases were checked in descending order of length, and the ruler of the first alias found as a substring of the auction description was taken as the ruler named in that description. The RIC number was then recovered with a regular expression that accepts common variants such as \texttt{RIC 123}, \texttt{R.I.C.\ 123}, and \texttt{RIC IV 123}, retaining only the numeric part. Where both a ruler and an RIC number were found, they were combined into a coarse catalogue type identifier such as \emph{Hadrian 123}; otherwise, the identifier was recorded as \texttt{unknown}. These identifiers allow coins belonging to the same catalogue type to be grouped, which is needed to train the bi-encoder introduced in Section~\ref{sec:reverse-descriptions}. This approach is deliberately simple; records that name a ruler absent from the list receive no identifier. We nonetheless chose exact string matching over more flexible methods because the identifiers are used to fine-tune the bi-encoder, where we prioritized data quality over quantity. The output of this stage is then a JSON object with the keys "obverse\_legend", "reverse\_legend", "obverse\_description", "reverse\_description" and "ric", for each record.

\begin{figure}[t]
\small
\begin{minipage}[t]{0.49\linewidth}
\textbf{(a) Raw auction description}
\begin{lstlisting}[style=jsonstyle, basicstyle=\ttfamily\scriptsize, breaklines=true, breakatwhitespace=true, breakindent=0pt]
MAXIMINUS II DAIA (310-313). BI Argenteus. Treveri.<br>Obv:  IMP MAXIMINVS AVG.<br>Radiate, draped, and cuirassed bust left, raising hand and holding globe.<br>Rev: SOLI INVICTO COMITI / P TR.<br>Sol driving facing quadriga, head left, raising hand and holding globe and whip.<br>RIC 826.<br>Condition:  Very fine.<br>Weight:    2.34 g.<br>Diameter:  19 mm.
\end{lstlisting}
\end{minipage}\hfill
\begin{minipage}[t]{0.49\linewidth}
\textbf{(b) Structured record}
\begin{lstlisting}[style=jsonstyle, basicstyle=\ttfamily\scriptsize]
{
  "obverse_legend": "IMPMAXIMINVSAVG",
  "reverse_legend": "SOLIINVICTOCOMITIPTR",
  "obverse_description": "Maximinus II; radiate bust facing left, draped and cuirassed, holding globe",
  "reverse_description": "Sol driving facing quadriga, head left, raising hand and holding globe and whip.",
  "ric": "Maximinus II 826"
}
\end{lstlisting}
\end{minipage}

\caption{Example auction description and structured labels extracted through our pipeline. The obverse description is shown after standardisation.}
\label{fig:extraction}
\end{figure}

\subsubsection{Image Processing}
Each image is converted to greyscale and smoothed with a $5 \times 5$ Gaussian filter using OpenCV~\citep{bradski2000opencv}, with $\sigma$ set to 0 so that it is derived from the kernel size. The image is then reduced to a horizontal profile by summing, for each column, the absolute horizontal derivative computed with a $3 \times 3$ Sobel operator~\citep{sobel1968}. The horizontal profile is smoothed with an 11-pixel moving average and normalized by its maximum. Columns with a value above 0.2 are marked as containing part of a coin, and runs of at least 10 consecutive marked columns are taken as candidate coin sides. When two or more runs are found, the first two from the left are taken as the obverse and reverse. When a single run is found, the image is split at the column of lowest energy within it, provided that column lies in the middle 50\% of the run's width (between one quarter and three quarters of the way across). Otherwise the run is split at its midpoint, since a minimum near either end is more likely to lie within a coin than in the gap between the two. Each crop is then bounded vertically using the analogous row-wise profile of the vertical derivative, with the same parameters, taking the span from the first to the last run plus a 2-pixel margin. Finally, each crop is taken from the original coloured image, so we recover the colour, and resized to $512 \times 512$ pixels using area interpolation.

\subsubsection{Standardizing Obverse Descriptions}\label{sec:standardising-obv-descs}
The obverse of a Roman Imperial coin almost always shows an imperial portrait with a small, recurring set of attributes, so it can be described in a regular way and captured by a fixed taxonomy. However, the auction descriptions in the original corpus do not follow this structure consistently and typically omit some information. Training directly on such text would therefore obscure the structure the model should be learning, so we standardized the obverse descriptions against a taxonomy derived from the standard classification of Roman bust types~\cite{pas_bust_types}. The taxonomy is organized into the following components:
\begin{itemize}
    \item \textbf{Portrait subject:} the name of the member of the imperial house depicted in the portrait, most often the emperor.

    \item \textbf{Portrait orientation:} left or right.  A \emph{frontal} class was initially considered but dropped from the taxonomy, since very few coins show a frontal portrait. Such coins are retained and assigned \emph{left} or \emph{right} by the orientation classifier described below, according to the side towards which the portrait leans; for the small number of fully frontal portraits this label is necessarily arbitrary.
    \item \textbf{Portrait type:} whether the portrait is a \emph{head}, which ends at the neck, or a \emph{bust}, which also shows the shoulders and upper chest, and therefore the clothing worn. Drapery may nonetheless be visible at the truncation of a head, so the dress component is recorded for both.

    \item \textbf{Headwear:} \emph{bare}, \emph{laureate}, \emph{radiate}, \emph{diademed},
    \emph{helmeted}, or \emph{veiled}.

    \item \textbf{Dress:} \emph{draped}, \emph{cuirassed},
    \emph{draped and cuirassed}, \emph{drapery on left shoulder}, \emph{consular robes}, or \emph{aegis}.

    \item \textbf{Attributes:} \emph{holding spear},
    \emph{holding spear and shield on shoulder}, \emph{holding sceptre},
    \emph{holding eagle-topped sceptre}, \emph{holding globe}, or \emph{holding Victory on globe}. This component is multi-label, since more than one attribute may be present.
\end{itemize}

Each component was first recovered deterministically from the auction description, where possible, by searching for terms associated with the corresponding taxonomy component. For example, portrait orientation was identified from expressions such as \emph{facing right} or \emph{facing left}. When none of the predefined values for a component could be recovered from the description, that component was instead predicted using a dedicated image classifier.

A separate vision-transformer classifier was trained for each component of the taxonomy. For the simpler components of portrait orientation, headwear, and portrait type, we used the \texttt{facebook/deit-small-patch16-224}\footnote{\url{https://huggingface.co/facebook/deit-small-patch16-224}}~\citep{touvron2021training} model; for the more difficult components of portrait subject, dress, and attributes, we used the \texttt{google/vit\-large-patch16-224}\footnote{\url{https://huggingface.co/google/vit-large-patch16-224}}~\citep{wu2020visual} model, which was pre-trained on ImageNet-21k and fine-tuned on ImageNet-1k~\citep{deng2009imagenet}. Each classifier was trained only on coins for which the corresponding label could be extracted deterministically from the auction description, and these were split into training, validation, and test sets in an 80/10/10 ratio, stratified by label. All parameters were fine-tuned using AdamW with a batch size of 32, weight decay of 0.01, and a linear learning-rate schedule with 20 warm-up steps, in bfloat16 precision. The checkpoint with the highest validation macro-F1 was retained (micro-F1 for the multi-label attribute classifier, whose predictions use a sigmoid threshold of 0.5). The learning rate and number of epochs for each classifier are given in Table~\ref{tab:obv-desc-classifiers}. After this process, each coin had a value for each of the taxonomy components. We then combined these values to generate an obverse description for each coin, following a standard template:
\begin{quote}
\emph{$\langle$portrait subject$\rangle$; $\langle$headwear$\rangle$
$\langle$portrait type$\rangle$ facing $\langle$portrait orientation$\rangle$,
$\langle$dress$\rangle$, $\langle$attributes$\rangle$},
\end{quote}

For example: \emph{Augustus; laureate head facing right, draped, holding spear}.

\subsubsection{Dataset Splits}
Filtering was performed separately for each task before the data were split. A record was excluded if the cropped image for the relevant coin side was unavailable or if the corresponding target field had been recorded as \texttt{unknown} during structured extraction. Since a record could, for example, contain a valid legend but an unknown description, the filtered sets differ between tasks, giving four separate datasets, one per task. Each dataset was then randomly divided into training, validation, and test sets in an 80/10/10 ratio. The validation set was used to monitor training, and the test set was held out for final evaluation, so all metrics reported in Section~\ref{sec:results} are computed on specimens unseen during training. Splits were drawn independently for each task, but since a separate adapter is trained for each, no coin is used for both training and evaluation of the same adapter. Table~\ref{tab:dataset-sizes} summarizes the resulting dataset sizes. Of the obverse- and reverse-legend test sets, 109 and 283 coins respectively carry a legend that does not occur in the corresponding training set; these form the unseen-legend subsets used in Section~\ref{sec:results}.

\begin{table}[t]
\centering
\small
\caption{Final dataset sizes after filtering and splitting.}
\label{tab:dataset-sizes}
\begin{tabular}{l
                S[table-format=5.0]
                S[table-format=4.0]
                S[table-format=4.0]
                S[table-format=5.0]}
\toprule
{Dataset} & {Train} & {Validation} & {Test} & {Total} \\
\midrule
Obverse legends      & 14735 & 1842 & 1842 & 18419 \\
Reverse legends      & 15436 & 1930 & 1930 & 19296 \\
Obverse descriptions & 24372 & 3047 & 3047 & 30466 \\
Reverse descriptions & 24361 & 3044 & 3046 & 30451 \\
\bottomrule
\end{tabular}
\end{table}

\subsection{Model Adaptation}\label{sec:training}

\subsubsection{Base Model and QLoRA}\label{sec:base-and-qlora}
All local model training and inference are performed on a single workstation equipped with one NVIDIA RTX A4500 GPU with 20\,GB of VRAM and 64\,GB of system RAM. The common backbone for all four tasks is Qwen3-VL-8B-Instruct~\citep{bai2025}, the instruction-tuned variant of the 8-billion-parameter Qwen3-VL model, obtained from Hugging Face in the pre-quantized 4-bit bitsandbytes release distributed by Unsloth (\texttt{unsloth/Qwen3-VL-8B-Instruct-unsloth-bnb-4bit})\footnote{\url{https://huggingface.co/unsloth/Qwen3-VL-8B-Instruct-unsloth-bnb-4bit}}. A separate set of adapters is trained for each of the four tasks using the Unsloth library~\citep{unsloth}. Adaptation follows the QLoRA framework~\citep{dettmers2023}: the quantized base weights are frozen, while low-rank adapters, with rank $r=8$, scaling factor $\alpha=16$, and no dropout, are inserted into all linear layers of the attention and feed-forward blocks of both the vision encoder and the language model; these adapters are the only trainable parameters. Training minimizes the standard autoregressive next-token cross-entropy loss over the text tokens of each training sequence. All four adapters are trained with the 8-bit AdamW optimizer~\citep{loshchilov2019adamw,dettmers2022optimizers}, using a per-device batch size of 2 with 4 gradient-accumulation steps (an effective batch size of 8), a peak learning rate of $5\times10^{-5}$ reached after 5 warm-up steps and decayed linearly, and a weight decay of 0.001. The adapter weights at the end of training are the ones used for evaluation.

\subsubsection{System Prompts}
Each task uses a fixed system prompt, which specifies the required output, together with a user message containing a short instruction and the coin image. Both legend-transcription tasks share a single system prompt (Appendix~\ref{appendix:legend-extraction-prompt}), with the user message ``Extract the legend from this image of a Roman coin''. The two description tasks each have their own system prompt (Appendices~\ref{appendix:obverse-description-extraction-prompt} and~\ref{appendix:reverse-description-extraction-prompt}), with the user message ``Extract the description from this image of a Roman coin''. The same system prompts and user messages are used during fine-tuning and inference, and for all models evaluated, including both baselines. Model outputs are canonicalized before scoring (Section~\ref{sec:metrics}), so differences in spacing, punctuation, or case do not affect evaluation.

\subsubsection{Legend Transcription Training}
Preliminary fine-tuning experiments on the legend-transcription tasks revealed a recurring failure mode. In some cases, the model generated a legend that was plausible for the depicted emperor but disagreed with characters that were clearly legible on the coin, suggesting that it was inferring a likely inscription from the central imagery rather than reading the characters around the rim. To encourage the model to attend to the rim, we generated an additional training image for each coin in which the central imagery is strongly blurred while the blur weakens towards the rim, where the legend is located. We did not train a model on the original images alone, so the contribution of this augmentation to the results in Section~\ref{sec:results} is not isolated; we present it as part of the training recipe rather than as a validated improvement.

Let $I$ denote the original $512 \times 512$ image and $B$ a copy blurred with a Gaussian filter of standard deviation $\sigma = 40$ pixels, applied using OpenCV~\citep{bradski2000opencv}. The augmented image $\hat{I}$ is constructed as

\begin{equation}
\hat{I}(x,y) = w(r)\,B(x,y) + \bigl(1-w(r)\bigr)\,I(x,y),
\qquad
w(r) = \exp\!\left(-\left(\frac{r}{R}\right)^{p}\right),
\end{equation}

where $r$ is the distance from pixel $(x,y)$ to the image centre, $R = 0.45\,\min(h, w)$ for an image of height $h$ and width $w$, and $p = 5$ controls how sharply the weight falls off. The weight equals 1 at the centre, where the blurred copy dominates, and decreases with radius, so the original image contributes increasingly towards the rim. Therefore, the transformation suppresses the portrait or central reverse design, while largely retaining the inscription. Each blurred image is assigned the same legend target as its original. The two legend-transcription adapters, one for the obverse and one for the reverse, are each trained for two epochs on the union of the original and blurred training images, so every training coin appears twice per epoch, once in each form, with the combined set reshuffled at the start of each epoch. At test time, models are evaluated on the original, unblurred images only.

\subsubsection{Description Generation Training}
The obverse- and reverse-description models use the same QLoRA configuration described in Section~\ref{sec:base-and-qlora} and are trained for two epochs, without image augmentation, using the standardized obverse descriptions of Section~\ref{sec:standardising-obv-descs} and the extracted reverse descriptions as targets.

\subsection{Evaluation}\label{sec:metrics}
\subsubsection{Baselines and Inference Protocol}
Each model is evaluated on the held-out test partition described in Section~\ref{sec:dataset}. Two general-purpose VLMs are used as baselines. The first is the same \texttt{unsloth/Qwen3-VL-8B-Instruct\-unsloth-bnb-4bit} backbone without the trained LoRA adapters; this comparison isolates the effect of domain-specific fine-tuning. The second is GPT-4.1 (\texttt{gpt-4.1-2025-04-14}), accessed through the OpenAI API, a proprietary, general-purpose VLM. The Qwen models, which we run locally, use greedy decoding, selecting the token to which the model assigns the highest probability at each step, with a maximum output length of 128 tokens. GPT-4.1 is queried with a temperature of zero. However, exact repeatability should not be assumed for GPT-4.1, since API-served models may return slightly different outputs for identical requests even at zero temperature, due to server-side implementation factors outside the caller's control~\citep{ouyang2025nondeterminism}.

For the two description tasks, we report the evaluation metrics on the full held-out test set. For the legend-transcription tasks, however, we report results on both the full test set and the \emph{unseen-legend subset} described in Section~\ref{sec:dataset}. The full test set measures overall performance, including on familiar inscriptions, while the unseen subset provides a stricter test of whether the model has learned to transcribe previously unseen legends rather than simply reproducing ones encountered during training.

\subsubsection{Legend Transcription}
Legend transcription is evaluated in the same way for the obverse and reverse sides. Let $g$ denote the ground-truth legend and $\hat{g}$ the legend generated by a model. Before scoring, both strings are canonicalized using the same normalization applied during dataset construction, giving $g'$ and $\hat{g}'$. Both metrics below are computed per legend and averaged over the test set or subset being evaluated. Since an exact transcription is the goal, but near misses should be distinguished from complete failures, the primary metric is the Levenshtein distance~\citep{levenshtein1966} normalized by the length of the longer string, which we refer to as the normalized Levenshtein distance (NLD):
\begin{equation}\label{eq:nld}
\mathrm{NLD}(g',\hat{g}') = \frac{d_{\mathrm{lev}}(g',\hat{g}')}{\max\bigl(|g'|,|\hat{g}'|\bigr)},
\end{equation}
where $d_{\mathrm{lev}}$ is the minimum number of single-character insertions, deletions, and substitutions required to transform $g'$ into $\hat{g}'$, and $|g'|$ and $|\hat{g}'|$ denote the lengths, in characters, of the ground-truth and generated legends, respectively.

NLD values lie in $[0,1]$, with $0$ corresponding to an exact transcription. Normalizing by the longer string makes errors comparable across inscriptions of different lengths. However, NLD treats errors at all positions equally, whereas in practice errors near the beginning of a legend are often more significant than errors in the middle, since a run of incorrect characters at the start is likely to indicate that the model has misread the legend entirely. We therefore also measure how accurately the model recovers the beginning of the legend using Prefix-5 accuracy (F5). With $k=\min(5,|g'|)$,
\begin{equation}\label{eq:prefix-acc}
\mathrm{F5}(g',\hat{g}') = \frac{1}{k}\sum_{i=1}^{k}\mathbb{I}\bigl[\, i\leq|\hat{g}'| \land \hat{g}'_{i}=g'_{i} \,\bigr],
\end{equation}
where $\mathbb{I}[\cdot]$ is the indicator function, so that a character position missing from a prediction shorter than the prefix is counted as incorrect. F5 values also lie in $[0,1]$, with $1$ indicating that all of the first five reference characters have been transcribed correctly.

\subsubsection{Reverse Descriptions}\label{sec:reverse-descriptions}
Reverse descriptions are open-ended. The same coin type can be described correctly using different vocabulary, word order, and levels of detail, making exact string comparison unsuitable. At the same time, semantic similarity alone is insufficient; a fluent description can capture the broad scene while omitting or misidentifying the figure, attributes, or objects that distinguish one reverse type from another. Therefore, we evaluate reverse descriptions using two complementary measures: a domain-adapted semantic-similarity score and a structured content score.

The semantic metric uses a domain-adapted bi-encoder. We fine-tune \texttt{BAAI/bge-large-en-v1.5}~\citep{xiao2024cpack} with the multiple-negatives ranking loss~\citep{reimers2019sbert}, treating two reverse descriptions as a positive pair if their coins share the same catalogue identifier from Section~\ref{sec:structured-label-extraction}. To ensure that the bi-encoder never sees the reference descriptions against which generated descriptions are later scored, all of its data are drawn from the training split of the reverse-description dataset. Within this split, descriptions sharing an identifier are de-duplicated and identifiers with fewer than two distinct descriptions are discarded. For each remaining identifier, 15\% of its descriptions, and at least two, are held out, unless this would leave fewer than two for training, in which case all of its descriptions are used for training. This yields 12,221 training descriptions from 3,901 identifiers, from which every pair of descriptions with the same identifier forms a positive pair (28,042 pairs in total), and 2,759 held-out descriptions from 1,342 identifiers. Since the loss treats the other descriptions in a batch as negatives, each batch of 16 pairs is drawn from 16 distinct identifiers, so that no in-batch negative is in fact a positive. The model is trained for one epoch, with a learning rate of $10^{-6}$ and 50 warm-up steps, and the weights at the end of the epoch are the ones used. To evaluate the encoder before and after fine-tuning, 500 positive pairs, 500 negative pairs, and 500 triplets are sampled once from the held-out descriptions with a fixed seed; the metrics computed on them are defined and reported in Section~\ref{sec:reverse-descriptions-results}.

To score a generated reverse description, $\hat{y}$, against its ground-truth $y$, both are mapped by the fine-tuned bi-encoder, $f$, to embedding vectors, and their cosine similarity is rescaled from $[-1,1]$ to $[0,1]$:
\begin{equation}
\mathrm{SemSim}(y,\hat{y}) = \frac{1}{2}\left(\frac{f(y)\cdot f(\hat{y})}{\lVert f(y)\rVert\,\lVert f(\hat{y})\rVert} + 1\right),
\end{equation}
where $\lVert\cdot\rVert$ is the Euclidean norm.

 The semantic similarity metric rewards descriptions that are close in overall meaning to the ground-truth description. However, it does not guarantee that specific numismatically important details, such as the name of the entity depicted, have been recovered. To measure whether these details are recovered, each reference description in the test set was first converted by the \texttt{gpt-4.1-mini} model~\citep{openai2025gpt41mini}, with a temperature of 0 and structured outputs, into a structured representation. This representation consists of a principal entity $e$, the main figure depicted, and a set of salient objects $O=\{o_1,\ldots,o_n\}$, as shown in Figure~\ref{fig:reverse-structured-example}. The model received the system prompt given in Appendix~\ref{appendix:reverse-description-entity-extraction-prompt}, and a user message containing the reverse description followed by the instruction ``Extract the structured output''.

\begin{figure}[t]
\centering
\begin{minipage}{0.99\linewidth}
\small
\begin{lstlisting}[style=jsonstyle, basicstyle=\ttfamily\scriptsize]
{
  "reverse_description": "Hercules, naked, standing right, resting on club set on rock, lion-skin beside club.",
  "main_entity": "Hercules",
  "main_objects": [
    "club",
    "rock",
    "lion-skin"
  ]
}
\end{lstlisting}
\end{minipage}
\caption{An example structured reverse description, with the main entity and objects used by the structured evaluation score.}
\label{fig:reverse-structured-example}
\end{figure}

Let $M(a,\hat{y})$ be an indicator equal to one when phrase $a$ occurs as a substring of $\hat{y}$ after both are lowercased and punctuation is replaced by spaces, and zero otherwise. Half of the score is awarded for recovering the principal entity, and the remaining half is shared equally among the extracted objects:

\begin{equation}\label{eq:structured-score}
\mathrm{StructuredScore}(\hat{y})
=
0.5\,M(e,\hat{y})
+
\begin{cases}
\displaystyle
\frac{0.5}{n}
\sum_{i=1}^{n} M(o_i,\hat{y}),
& n>0,
\\[10pt]
0.5,
& n=0.
\end{cases}
\end{equation}

When no salient objects are extracted ($n=0$), the object component receives its full value of $0.5$. When no principal entity is extracted, the entity component receives $0$, which applies to only 73 of the 3,046 test descriptions.

\subsubsection{Obverse Descriptions}
Obverse descriptions differ from reverse descriptions because they have been standardized into the six-component taxonomy introduced in Section~\ref{sec:standardising-obv-descs}. They can therefore be evaluated component by component rather than by free-text similarity.

Both the generated description and its reference are first parsed into the six components. The text before the first semicolon is taken as the portrait subject, and the remaining components are recovered from the rest of the description with the same term-matching rules used to label the training data (Section~\ref{sec:standardising-obv-descs}). A component whose value cannot be identified, because no valid term or more than one conflicting term is present, is scored as incorrect, so a model that departs from the required format or vocabulary is penalized on the affected components.

Five components (portrait subject, portrait orientation, portrait type, headwear, and dress) are single-valued and are scored using binary exact match:
\begin{equation}
s_{c}^{i}
=
\begin{cases}
1, & \text{if the predicted value for component $c$ matches the reference for coin $i$,}\\
0, & \text{otherwise.}
\end{cases}
\end{equation}

The attributes component is multi-label, since a portrait may show zero or more attributes. Let $A_i$ and $\hat{A}_i$ denote the reference and predicted attribute sets for coin $i$. The component is scored with the set-level F1 score:
\begin{equation}
s_{\mathrm{attributes}}^{i} =
\begin{cases}
\dfrac{2\,|A_i \cap \hat{A}_i|}{|A_i| + |\hat{A}_i|}
  & \text{if } A_i \cup \hat{A}_i \neq \emptyset,\\[1.4ex]
1 & \text{otherwise.}
\end{cases}
\end{equation}
When exactly one of the two sets is empty the intersection is empty and the score is 0; when both are empty the model has correctly predicted that no attribute is present, and the score is set to 1.

The overall obverse-description score for coin $i$ is the unweighted mean of the six component scores:

\begin{equation}\label{eq:obverse-score}
S_i
=
\frac{
s_{\mathrm{subject}}^{i}
+
s_{\mathrm{orientation}}^{i}
+
s_{\mathrm{type}}^{i}
+
s_{\mathrm{headwear}}^{i}
+
s_{\mathrm{dress}}^{i}
+
s_{\mathrm{attributes}}^{i}
}{6}.
\end{equation}

In Section~\ref{sec:results}, we report the mean overall score and the mean score on each component.

\section{Results}\label{sec:results}
\subsection{Legends}
For the legend transcription tasks, results are reported using the NLD and F5 metrics defined in Section~\ref{sec:metrics}, on both the full test set and the unseen-legend subset. The scatter plots provide a per-sample comparison of the NLDs achieved by two models. Each point corresponds to a single sample in the test set, the axes give the NLD achieved by each model on that coin, and the dashed diagonal marks equal performance, so points below the diagonal are samples where the model on the vertical axis performed better. Histograms show the distribution of F5 scores as percentages in 20\% bins. The more accurate the model, the more its F5 mass is concentrated in the higher bins.

\subsubsection{Obverse Legends}
\begin{figure}[t]
    \centering
    \begin{subfigure}[b]{0.85\linewidth}
        \centering
        \includegraphics[width=\linewidth]{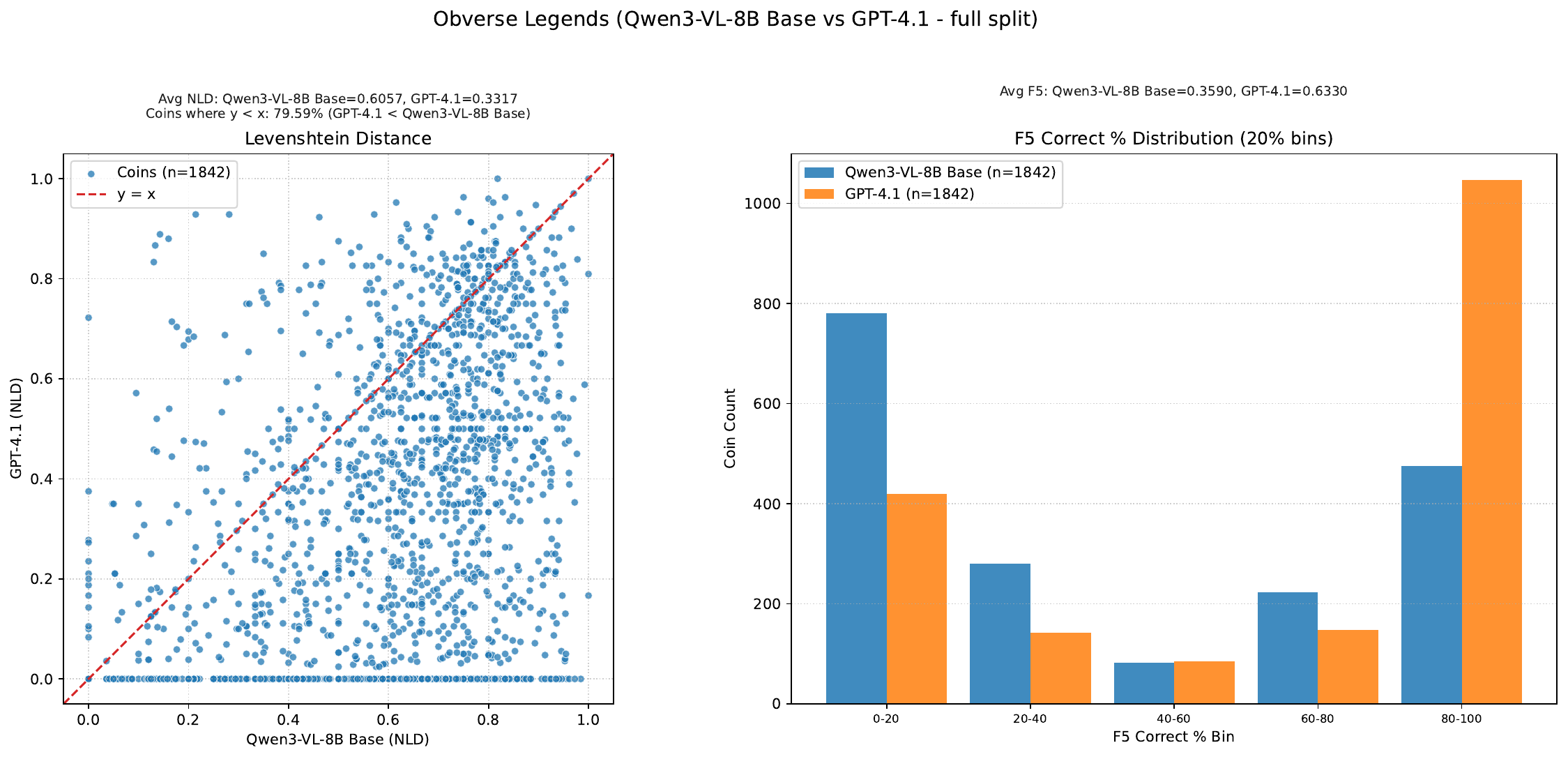}
        \caption{Full test set}
        \label{fig:obv-lgd-baselines-full}
    \end{subfigure}
    \hfill
    \begin{subfigure}[b]{0.85\linewidth}
        \centering
        \includegraphics[width=\linewidth]{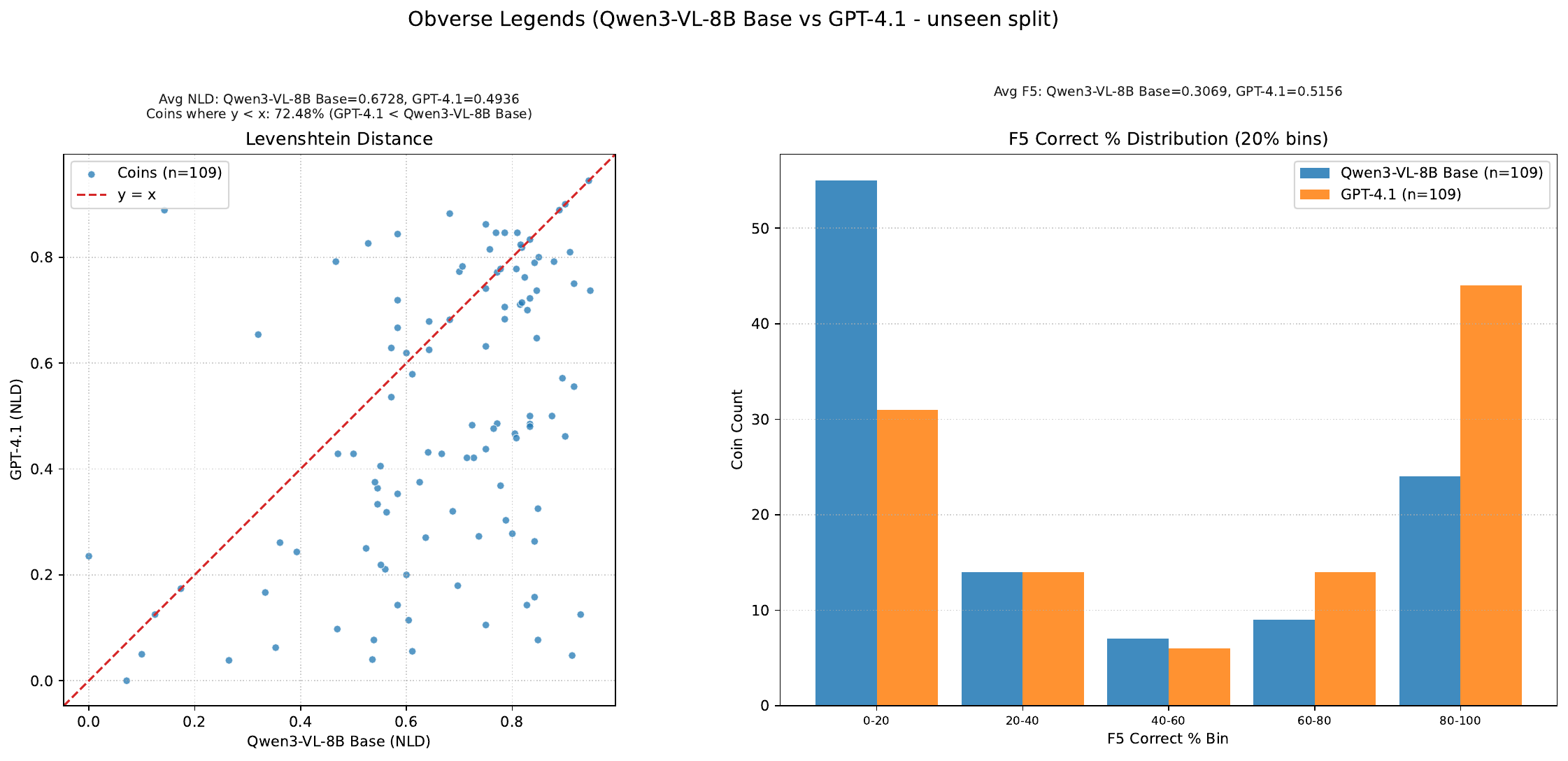}
        \caption{Unseen-legend subset}
        \label{fig:obv-lgd-baselines-unseen}
    \end{subfigure}
    \caption{Per-sample comparison of GPT-4.1 and the Qwen base model on obverse legend transcription.}
    \label{fig:obv-lgd-baselines}
\end{figure}

Figure~\ref{fig:obv-lgd-baselines} compares the two baselines on the full test set and on the unseen-legend subset. GPT-4.1 clearly outperforms the unadapted Qwen base model, with most points lying below the diagonal and the F5 distribution shifted towards the higher bins. Neither baseline was trained on our data, so the seen/unseen distinction does not reflect memorization for them; the subset is reported here to provide a like-for-like reference for the fine-tuned model evaluated below.

\begin{figure}[t]
    \centering
    \begin{subfigure}[b]{0.85\linewidth}
        \centering
        \includegraphics[width=\linewidth]{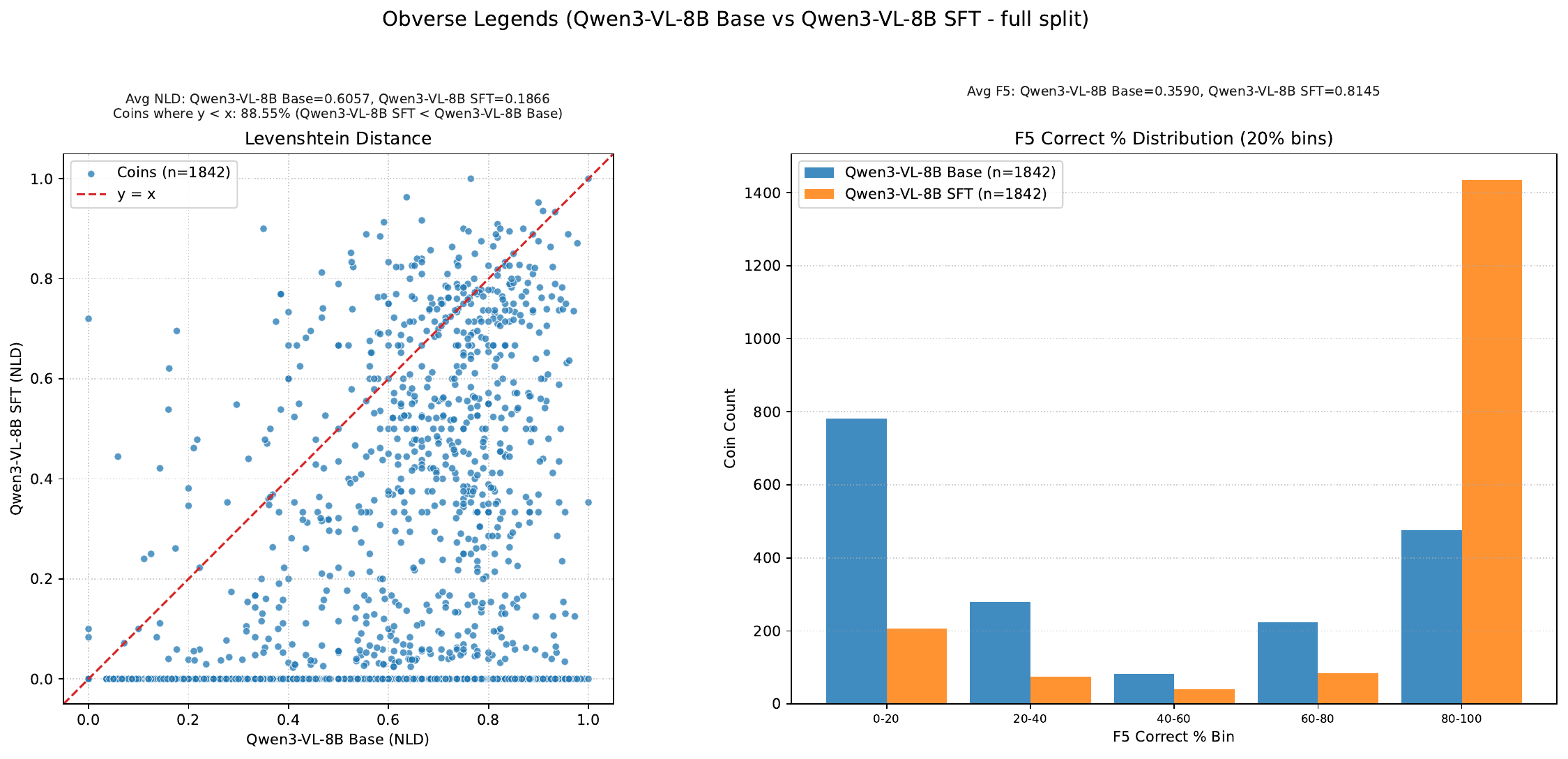}
        \caption{Full test set}
        \label{fig:obv-lgd-sft-vs-base-full}
    \end{subfigure}
    \hfill
    \begin{subfigure}[b]{0.85\linewidth}
        \centering
        \includegraphics[width=\linewidth]{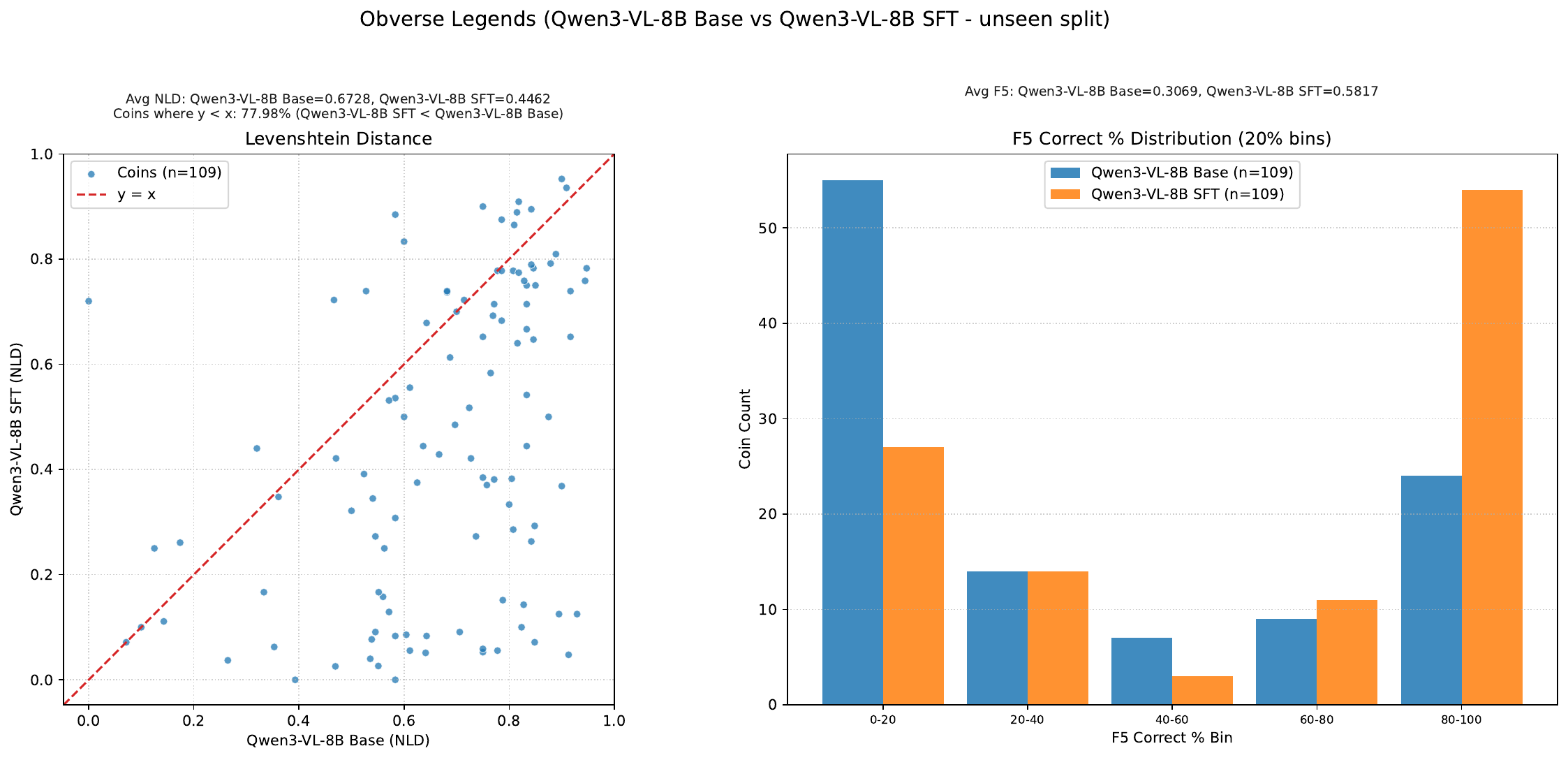}
        \caption{Unseen-legend subset}
        \label{fig:obv-lgd-sft-vs-base-unseen}
    \end{subfigure}
    \caption{Per-sample comparison of the fine-tuned model and the Qwen base model on obverse legend transcription.}
    \label{fig:obv-lgd-sft-vs-base}
\end{figure}

Figure~\ref{fig:obv-lgd-sft-vs-base} presents the corresponding comparison between the fine-tuned model and the Qwen base model. The effect of adaptation is clearly visible: the point cloud collapses towards the horizontal axis, with a dense band at $y=0$ corresponding to legends transcribed exactly by the fine-tuned model, and the F5 mass of the fine-tuned model is shifted towards the higher bins relative to the base model. The improvement is smaller on the unseen subset, but still clearly present, with most points remaining below the diagonal in the scatter plot.

\begin{table}[t]
\centering
\footnotesize
\setlength{\tabcolsep}{4pt}
\begin{minipage}[t]{0.49\linewidth}
\centering
\caption{Obverse legend transcription results. Lower NLD is better; higher F5 is better.}
\label{tab:obv-legends-results}
\begin{tabular}{ll
                S[table-format=1.2]
                S[table-format=1.2]}
\toprule
{Split} & {Model} & {NLD $\downarrow$} & {F5 $\uparrow$} \\
\midrule
\multirow{3}{*}{Full}
& Qwen base model & {0.61} & {0.36} \\
& GPT-4.1       & {0.33} & {0.63} \\
& Qwen with SFT & {\bfseries 0.19} & {\bfseries 0.81} \\
\midrule
\multirow{3}{*}{Unseen}
& Qwen base model & {0.67} & {0.31} \\
& GPT-4.1       & {0.49} & {0.52} \\
& Qwen with SFT & {\bfseries 0.45} & {\bfseries 0.58} \\
\bottomrule
\end{tabular}
\end{minipage}\hfill
\begin{minipage}[t]{0.49\linewidth}
\centering
\caption{Reverse legend transcription results. Lower NLD is better; higher F5 is better.}
\label{tab:rev-legends-results}
\begin{tabular}{ll
                S[table-format=1.2]
                S[table-format=1.2]}
\toprule
{Split} & {Model} & {NLD $\downarrow$} & {F5 $\uparrow$} \\
\midrule
\multirow{3}{*}{Full}
& Qwen base model & {0.65} & {0.30} \\
& GPT-4.1       & {0.50} & {0.50} \\
& Qwen with SFT & {\bfseries 0.26} & {\bfseries 0.77} \\
\midrule
\multirow{3}{*}{Unseen}
& Qwen base model & {0.68} & {0.27} \\
& GPT-4.1       & {0.58} & {0.48} \\
& Qwen with SFT & {\bfseries 0.45} & {\bfseries 0.62} \\
\bottomrule
\end{tabular}
\end{minipage}
\end{table}

Table~\ref{tab:obv-legends-results} reports the full results. The fine-tuned model outperforms both baselines on both metrics and on both splits. On the full test set, it reduces the NLD from 0.33 for GPT-4.1 to 0.19, while raising the F5 score from 0.63 to 0.81. The more demanding comparison is the unseen subset, where every target legend is absent from the training set; here the fine-tuned model still improves on GPT-4.1, reducing the distance from 0.49 to 0.45 and raising the F5 score from 0.52 to 0.58. Therefore, fine-tuning does more than make the model better at recalling legends it has seen during training: the gain on inscriptions the model has never seen as whole legends indicates that it has learned a transferable ability to read characters from the coin surface. However, the gap between the two splits shows that part of the full-set performance comes from familiarity with common legends, and the model remains noticeably better at extracting legends contained in the training set.

\subsubsection{Reverse Legends}
\begin{figure}[t]
    \centering
    \begin{subfigure}[b]{0.85\linewidth}
        \centering
        \includegraphics[width=\linewidth]{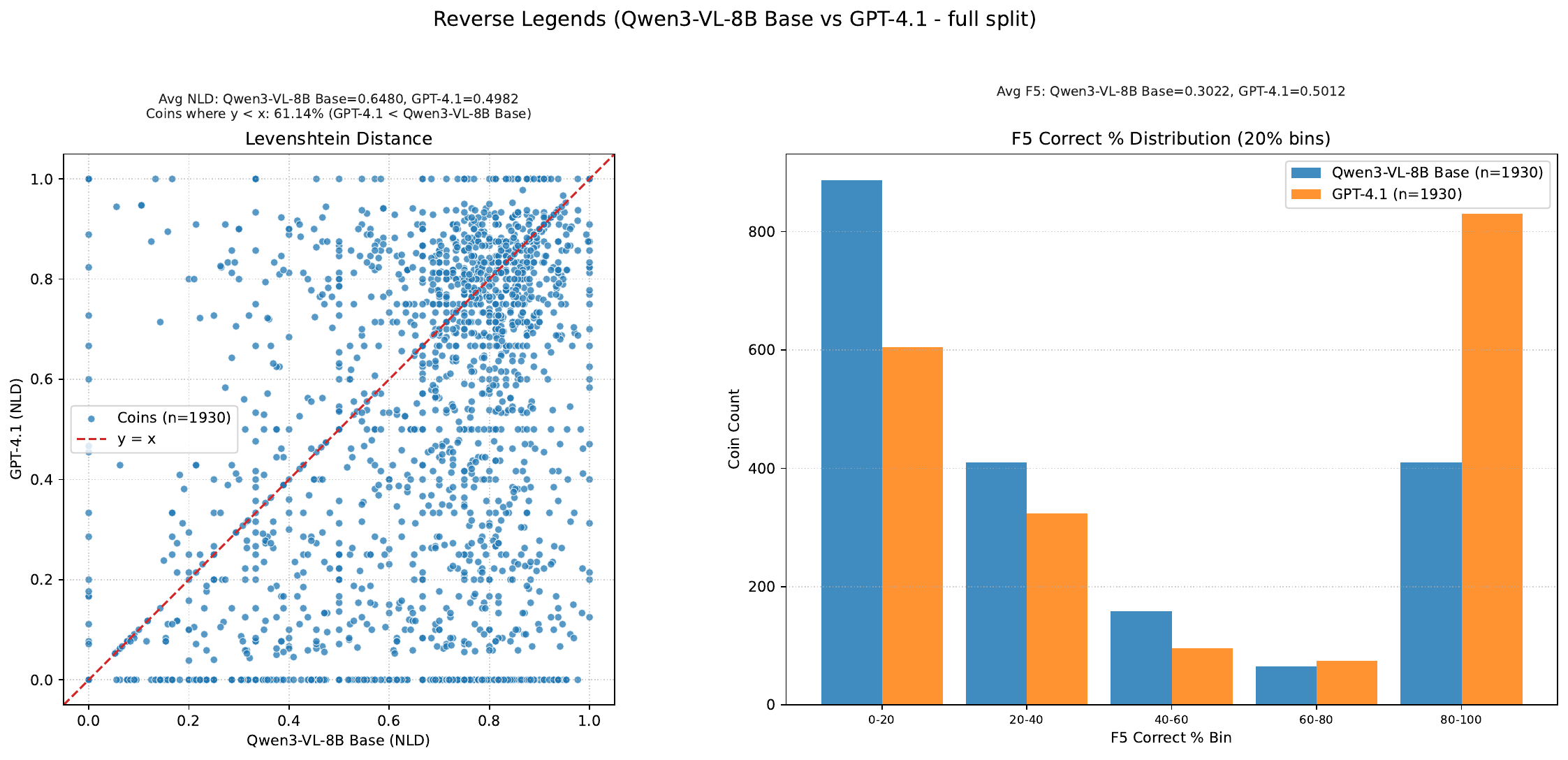}
        \caption{Full test set}
        \label{fig:rev-lgd-baselines-full}
    \end{subfigure}
    \hfill
    \begin{subfigure}[b]{0.85\linewidth}
        \centering
        \includegraphics[width=\linewidth]{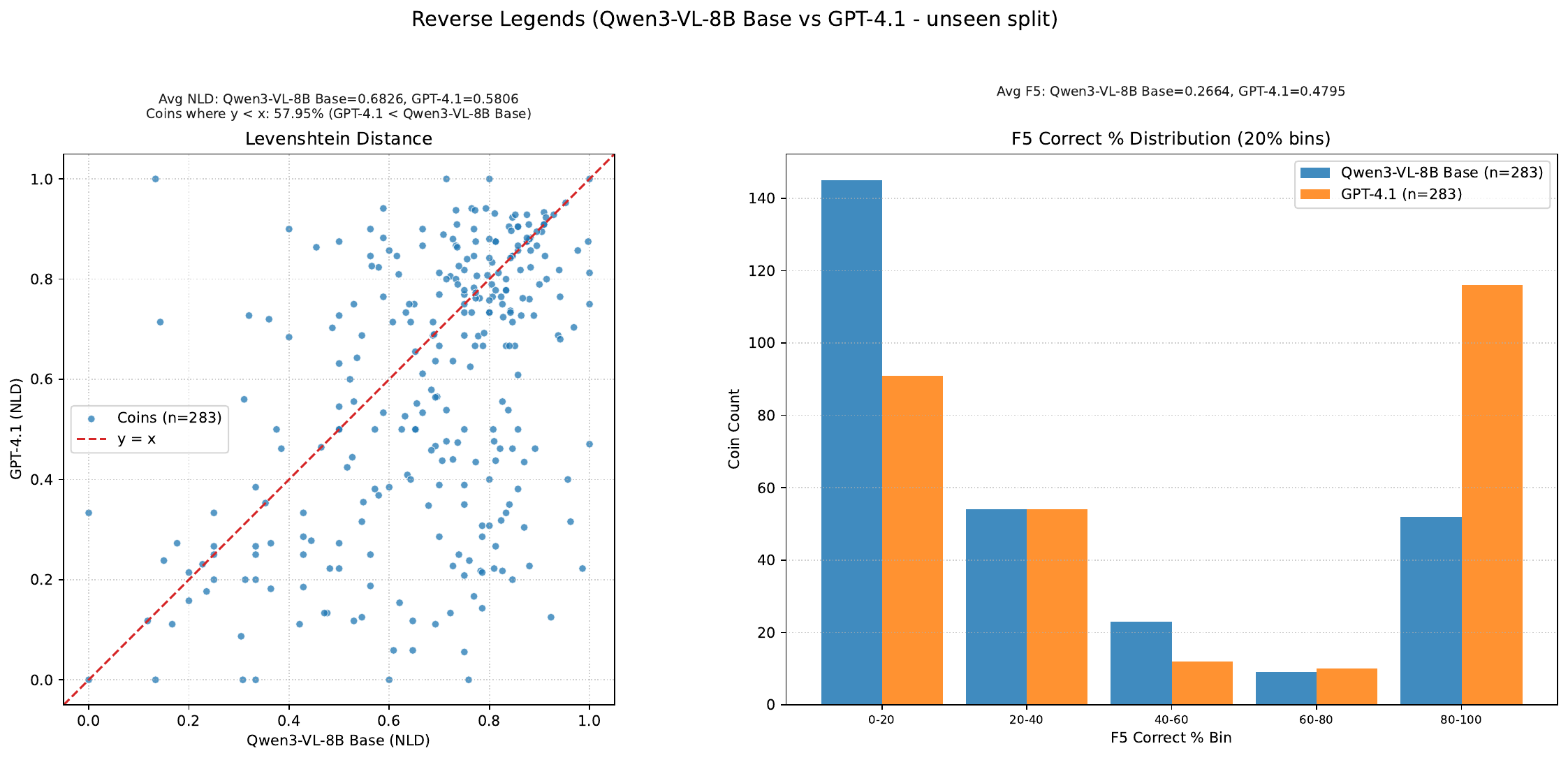}
        \caption{Unseen-legend subset}
        \label{fig:rev-lgd-baselines-unseen}
    \end{subfigure}
    \caption{Per-sample comparison of GPT-4.1 and the Qwen base model on
    reverse legend transcription.}
    \label{fig:rev-lgd-baselines}
\end{figure}

Figure~\ref{fig:rev-lgd-baselines} compares the two baselines on the full test set and on the unseen-legend subset. The pattern mirrors the obverse task, with GPT-4.1 clearly outperforming the unadapted Qwen base model on both metrics. However, both baselines perform worse on this task than on obverse legend transcription, suggesting that this is an intrinsically harder task, which reflects the greater variety of reverse legends.

\begin{figure}[t]
    \centering
    \begin{subfigure}[b]{0.85\linewidth}
        \centering
        \includegraphics[width=\linewidth]{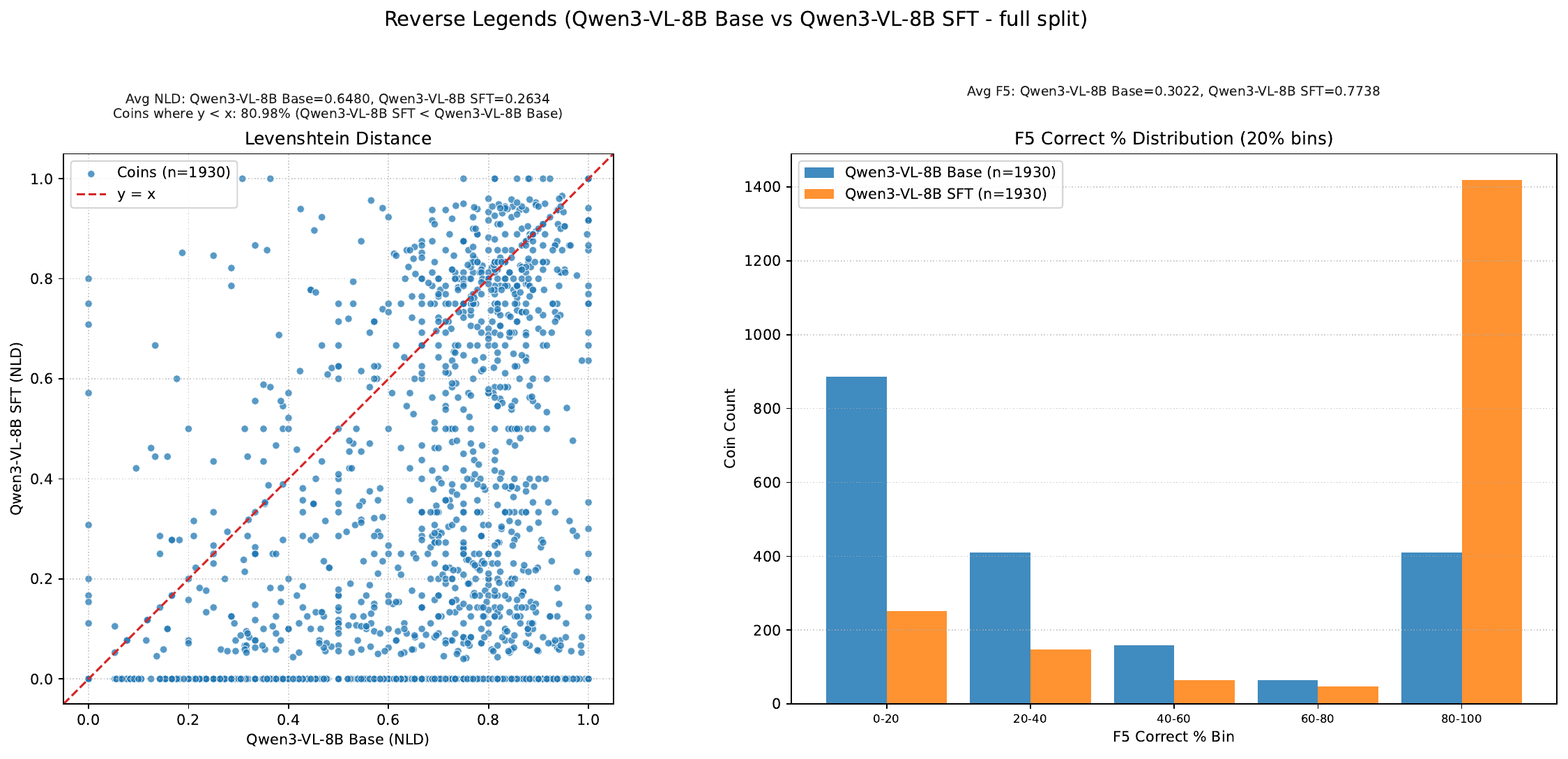}
        \caption{Full test set}
        \label{fig:rev-lgd-sft-vs-base-full}
    \end{subfigure}
    \hfill
    \begin{subfigure}[b]{0.85\linewidth}
        \centering
        \includegraphics[width=\linewidth]{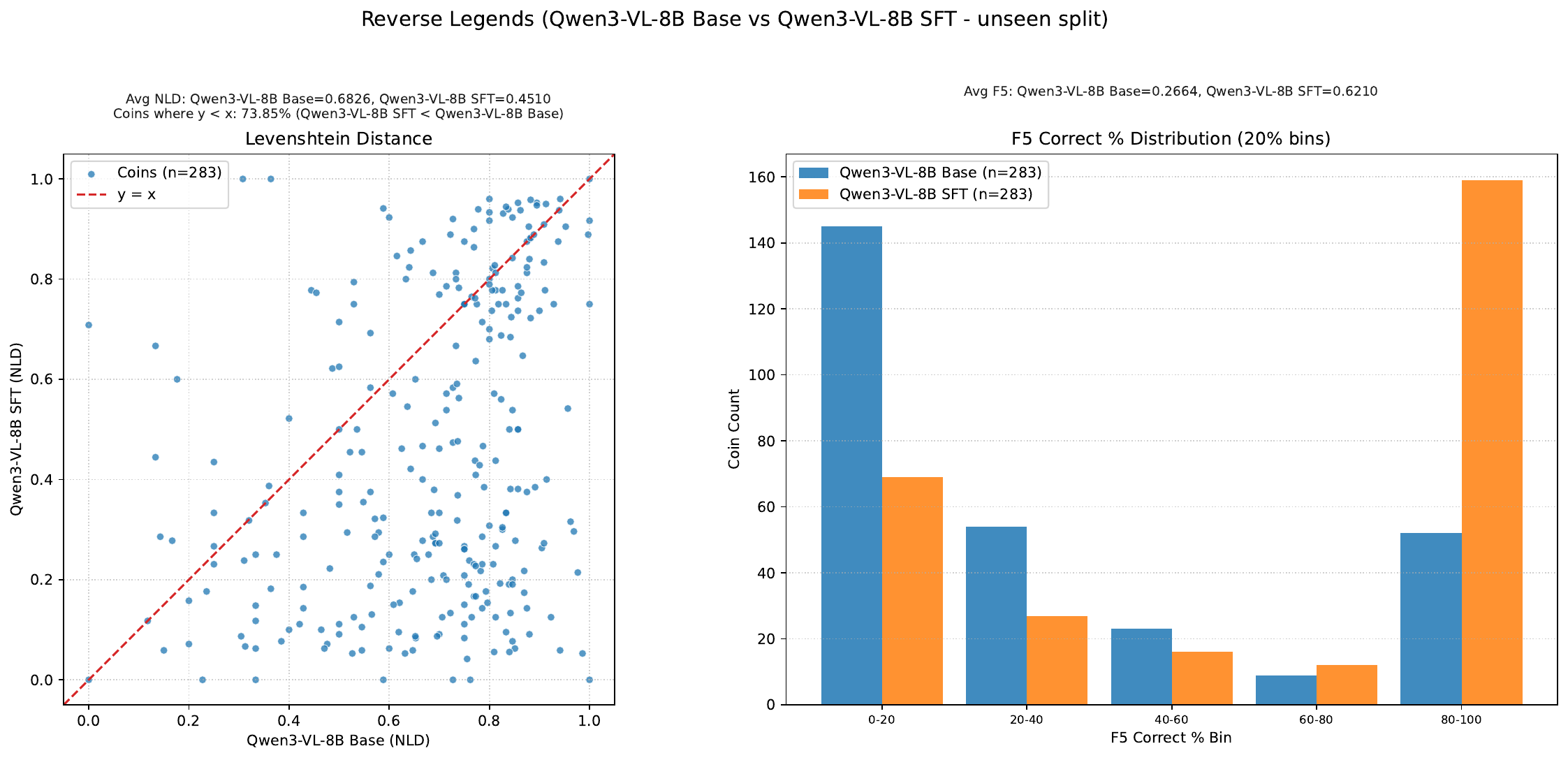}
        \caption{Unseen-legend subset}
        \label{fig:rev-lgd-sft-vs-base-unseen}
    \end{subfigure}
    \caption{Per-sample comparison of the fine-tuned model and the Qwen
    base model on reverse legend transcription.}
    \label{fig:rev-lgd-sft-vs-base}
\end{figure}

Figure~\ref{fig:rev-lgd-sft-vs-base} presents the comparison between the fine-tuned model and the Qwen base model. The pattern is the same as for obverse legends: the point cloud shifts firmly below the diagonal, and the F5 mass moves into the higher bins on both splits.

Table~\ref{tab:rev-legends-results} reports all scores. Again, the fine-tuned model outperforms both baselines on both metrics and on both splits. On the full test set it reduces the NLD from 0.50 for GPT-4.1 to 0.26, while raising the F5 score from 0.50 to 0.77. On the unseen subset the distance falls from 0.58 to 0.45 and the F5 score rises from 0.48 to 0.62. The generalization observed for obverse legends therefore holds for reverse legends as well, and the absolute improvement over GPT-4.1 is larger here.

\subsection{Descriptions}
\subsubsection{Reverse Descriptions}\label{sec:reverse-descriptions-results}
Reverse-description results are reported using the semantic similarity and structured score metrics introduced in Section~\ref{sec:metrics}. Since the semantic similarity metric calculation relies on the fine-tuned bi-encoder, we first report its performance before and after fine-tuning. All four metrics below are computed on held-out description pairs, with the similarity of a pair defined as the cosine similarity between the embeddings of the two descriptions:
\begin{itemize}
    \item \textbf{Triplet accuracy:} the proportion of triplets $(x, y, z)$, where $x$ and $y$ describe coins of the same catalogue type and $z$ a coin of a different type, in which $x$ is more similar to $y$ than to $z$.
    \item \textbf{ROC-AUC:} the area under the receiver operating characteristic curve when a similarity threshold is used to predict whether two descriptions refer to the same catalogue type.
    \item \textbf{PR-AUC:} the area under the precision--recall curve for the same binary classification task.
    \item \textbf{Similarity gap:} the difference between the mean similarity of same-type pairs and that of different-type pairs
\end{itemize}

The performance of the encoder before and after fine-tuning is given in Table~\ref{tab:rev-dsc-biencoder}; fine-tuning improves the score on every metric.

\begin{table}[t]
\centering
\small
\caption{Bi-encoder performance before and after fine-tuning}
\label{tab:rev-dsc-biencoder}
\begin{tabular}{
    l
    S[table-format=1.2, detect-weight=true]
    S[table-format=1.2, detect-weight=true]
    S[table-format=1.2, detect-weight=true]
    S[table-format=1.2, detect-weight=true]
}
\toprule
{Model}
& {\makecell{Triplet\\Acc.}}
& {\makecell{ROC-AUC}}
& {\makecell{PR-AUC}}
& {\makecell{Sim.\\Gap}} \\
\midrule
Base bi-encoder
& {0.86}
& {0.91}
& {0.93}
& {0.13} \\
Fine-tuned bi-encoder
& {\bfseries 0.89}
& {\bfseries 0.93}
& {\bfseries 0.94}
& {\bfseries 0.17} \\
\bottomrule
\end{tabular}
\end{table}

Having established that the fine-tuned bi-encoder separates same-type from different-type descriptions more cleanly than the base encoder, we use it to compute semantic similarity for the reverse-description results that follow. Each model comparison is shown in two panels. The left panel is a per-coin scatter plot of semantic similarity, with the Qwen base model on the horizontal axis and the comparison model on the vertical axis, so points above the dashed diagonal are coins on which the comparison model scored higher. The right panel shows the distribution of structured scores in 5\% bins; better-performing models place more mass in the higher bins.

\begin{figure}[t]
    \centering
    \includegraphics[width=0.85\linewidth]{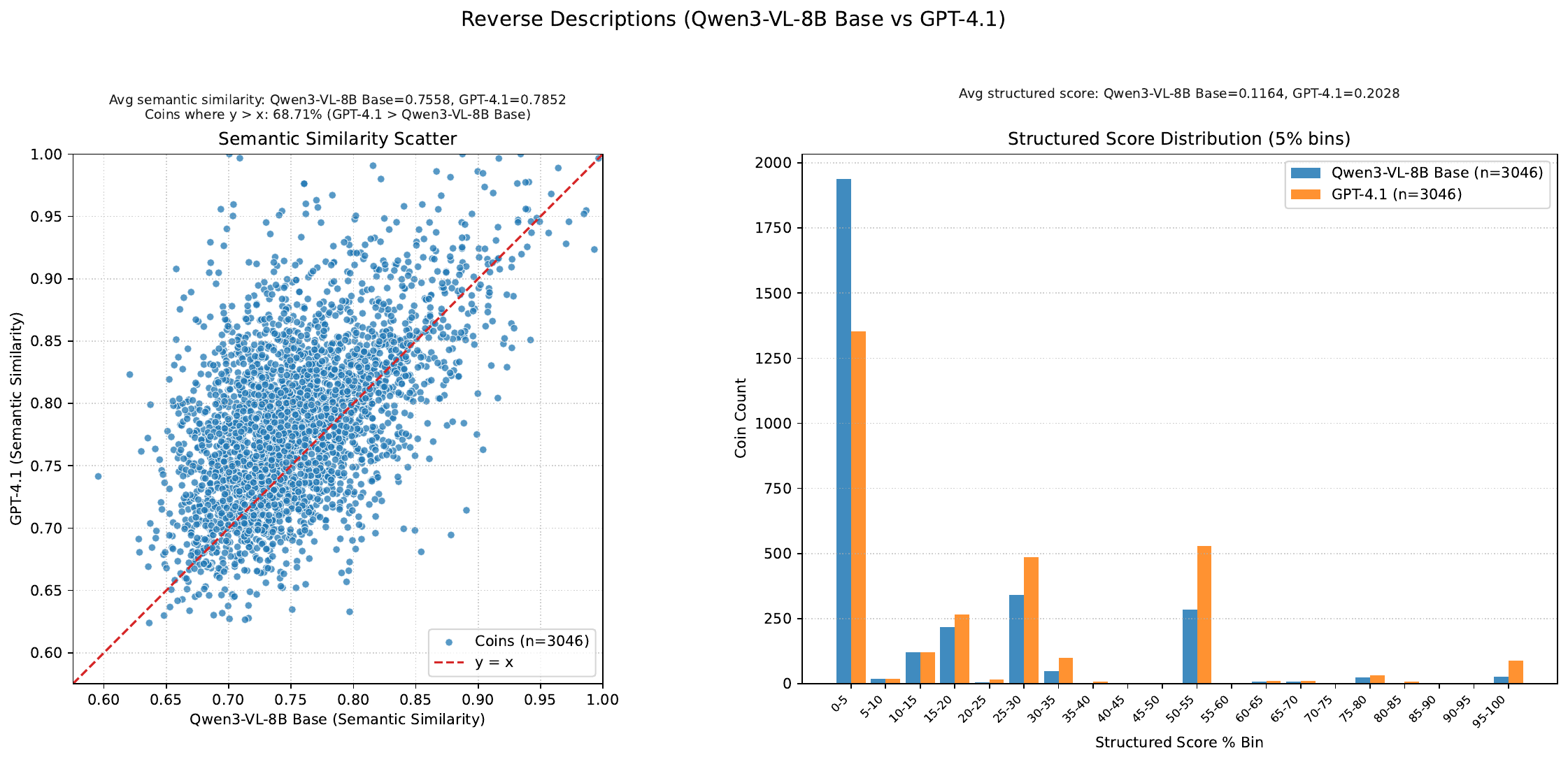}
    \caption{Comparison of GPT-4.1 and the Qwen base model on reverse-description generation. Left: per-sample semantic similarity. Right: distribution of structured scores.}
    \label{fig:rev-dsc-baselines}
\end{figure}

Figure~\ref{fig:rev-dsc-baselines} compares the two baselines: the Qwen base model and GPT-4.1. GPT-4.1 performs better on both metrics, but its structured score is poor in absolute terms.

\begin{figure}[t]
    \centering
    \includegraphics[width=0.85\linewidth]{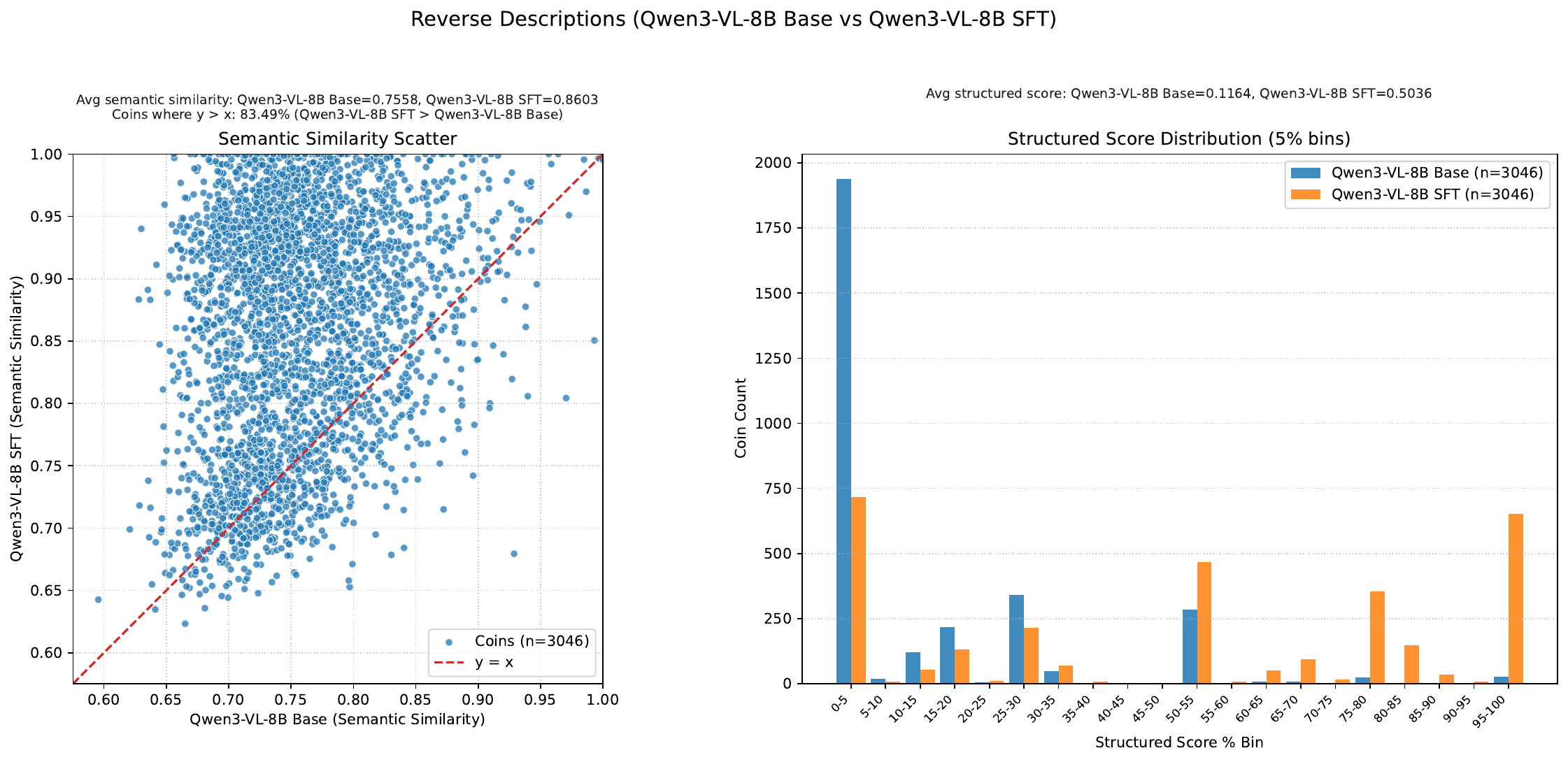}
    \caption{Comparison of the fine-tuned Qwen model and the Qwen base model on reverse-description generation. Left: per-sample semantic similarity.
    Right: distribution of structured scores.}
    \label{fig:rev-dsc-sft-vs-base}
\end{figure}

Figure~\ref{fig:rev-dsc-sft-vs-base} presents the corresponding comparison between the fine-tuned Qwen model and the Qwen base model. Table~\ref{tab:rev-dsc-results} reports all scores. 

\begin{table}[t]
\centering
\small
\caption{Reverse-description generation results.}
\label{tab:rev-dsc-results}
\begin{tabular}{l
                S[table-format=1.2]
                S[table-format=1.2]}
\toprule
{Model} & {\makecell{Semantic Similarity $\uparrow$}} & {\makecell{Structured Score $\uparrow$}} \\
\midrule
Qwen base model    & {0.76} & {0.12} \\
GPT-4.1          & {0.79} & {0.20} \\
Qwen with SFT    & {\bfseries 0.86} & {\bfseries 0.50} \\
\bottomrule
\end{tabular}
\end{table}

The fine-tuned model outperforms both the base model and GPT-4.1 on both metrics. Semantic similarity rises from 0.79 for GPT-4.1 to 0.86, and the structured score rises from 0.20 to 0.50. The contrast between the two metrics is informative: the semantic similarity improvement is smaller in both absolute and relative terms, since describing the general content of an image is already a strength of general-purpose VLMs. By contrast, the structured score improves by a factor of 2.5, suggesting that fine-tuning primarily adds the domain-specific vocabulary and specificity that general-purpose models lack.

\subsubsection{Obverse Descriptions}
As described in Section~\ref{sec:standardising-obv-descs}, dedicated classifiers were used to fill taxonomy components that could not be recovered deterministically from the auction descriptions. We first report the performance of these classifiers. Table~\ref{tab:obv-desc-classifiers} gives, for each classifier: the size of the dataset used to train and evaluate it, its architecture, and its performance on the test set. Portrait orientation and headwear are predicted most accurately, while portrait subject and dress are predicted considerably less accurately. The gap between accuracy and macro-F1 for dress (0.81 against 0.64) and, to a lesser extent, portrait subject (0.79 against 0.72) reflects the class imbalance in the labels. Both classifiers handle common classes well but perform noticeably worse on rare ones, such as consular robes among the dress labels. These errors propagate into the obverse descriptions the classifiers were used to complete, and therefore into both the training targets and the reference labels against which the VLMs are evaluated.

\begin{table*}[t]
\centering
\small
\caption{Training settings and test performance of the classifiers used to complete missing obverse-description taxonomy components.}
\label{tab:obv-desc-classifiers}
\begin{tabular}{lllrlccl}
\toprule
{Component} & {Task} & {Backbone} & {Classes} &
{N (train/val/test)} & {LR} & {Epochs} & {Test result} \\
\midrule
Portrait subject
& Multiclass
& ViT-Large
& 106
& 26,337 (21,092/2,622/2,623)
& $5\times10^{-5}$
& 8
& Acc 0.79, Macro-F1 0.72 \\

Portrait orientation
& Binary
& DeiT-Small
& 2
& 26,686 (21,348/2,669/2,669)
& $1\times10^{-5}$
& 1
& Acc 0.97, Macro-F1 0.90 \\

Portrait type
& Binary
& DeiT-Small
& 2
& 27,866 (22,292/2,787/2,787)
& $5\times10^{-5}$
& 1
& Acc 0.93, Macro-F1 0.92 \\

Headwear
& Multiclass
& DeiT-Small
& 6
& 24,095 (19,276/2,409/2,410)
& $5\times10^{-5}$
& 4
& Acc 0.95, Macro-F1 0.88 \\

Dress
& Multiclass
& ViT-Large
& 6
& 16,258 (13,006/1,626/1,626)
& $1\times10^{-5}$
& 8
& Acc 0.81, Macro-F1 0.64 \\

Attributes
& Multi-label
& ViT-Large
& 6
& 969 (781/94/94)
& $1\times10^{-5}$
& 8
& Micro/Macro F1 0.79/0.41 \\
\bottomrule
\end{tabular}
\end{table*}

We next report the performance of the VLMs on the obverse-description generation task. Table~\ref{tab:obv-dsc-results} reports, for each model, the mean score on each of the six taxonomy components together with the overall score. 

GPT-4.1 outperforms the Qwen base model on every component except portrait orientation, where both models score 0.97, raising the overall score from 0.51 to 0.65. The Qwen base model is particularly weak at portrait subject identification, correctly identifying fewer than 4\% of specimens. The fine-tuned model outperforms both GPT-4.1 and the Qwen base model on every component, and by a substantial margin on all but portrait orientation, raising the overall score to 0.81. The largest gain, in absolute and relative terms, is on portrait subject, which rises from 0.03 for the base model and 0.38 for GPT-4.1 to 0.68 for the fine-tuned model.

\begin{table*}[t]
\centering
\small
\caption{Obverse-description generation results. Higher is better for all columns.}
\label{tab:obv-dsc-results}
\begin{tabular}{
    l
    S[table-format=1.2, detect-weight=true]
    S[table-format=1.2, detect-weight=true]
    S[table-format=1.2, detect-weight=true]
    S[table-format=1.2, detect-weight=true]
    S[table-format=1.2, detect-weight=true]
    S[table-format=1.2, detect-weight=true]
    S[table-format=1.2, detect-weight=true]
}
\toprule
{Model}
& {Portrait subject}
& {Portrait Orientation}
& {Portrait type}
& {Headwear}
& {Dress}
& {\makecell{Attr.\\F1}}
& {\makecell{Overall\\Score}} \\
\midrule
Qwen base model
& {0.03}
& {0.97}
& {0.52}
& {0.65}
& {0.50}
& {0.37}
& {0.51} \\

GPT-4.1
& {0.38}
& {0.97}
& {0.75}
& {0.80}
& {0.64}
& {0.38}
& {0.65} \\

Qwen with SFT
& {\bfseries{0.68}}
& {\bfseries{0.99}}
& {\bfseries{0.91}}
& {\bfseries{0.92}}
& {\bfseries{0.76}}
& {\bfseries{0.58}}
& {\bfseries{0.81}} \\
\bottomrule
\end{tabular}
\end{table*}

\section{Discussion}\label{sec:discussion}
\subsection{Main Findings and Practical Implications}
This work shows that a general-purpose VLM can be adapted to the task of describing Roman Imperial coins through parameter-efficient fine-tuning on structured training data derived from auction records. Across all four tasks, the adapted models improve substantially over the unadapted base model and also outperform GPT-4.1, a proprietary general-purpose model. The adapted models also reach a useful level of absolute performance, with a mean normalized Levenshtein distance of 0.19 for obverse legends.

Fine-tuning and inference require only a single 20\,GB GPU through QLoRA, making domain-specific VLMs potentially accessible to museums, archives, and researchers. Although the models presented in this paper do not yet allow fully automated cataloguing of Roman Imperial coins, they could already be used to produce draft descriptions for experts to review.

\subsection{Generalization Beyond the Training Legends}
An important result is that the gains in legend transcription, relative to the baseline models, are not limited to inscriptions that occur in the training data. Fine-tuning improves performance on the unseen-legend subset, in which none of the ground-truth inscriptions is present in the training set, although the margin over GPT-4.1 is smaller there than on the full test set.

This distinction is central to the generative formulation proposed in this work. If improvements were restricted to previously seen legends, the model could in effect be behaving as a classifier implemented through autoregressive generation, with the legends observed during training forming an implicit label set. Instead, the improvement on unseen legends provides evidence that fine-tuning improves the model's underlying ability to transcribe coin inscriptions.

This property has practical implications. Legends vary even within a single coin type: abbreviations differ between dies, numerals in an emperor's titles change from year to year, and engravers' errors produce further variants, so a training set will rarely contain every legend later encountered. Worn coins, moreover, often preserve only part of a legend, and a partial transcription is a useful output that a predefined label is unlikely to contain. Similarly, a previously uncatalogued coin could, in principle, still have its legend transcribed. In each of these cases, a generative model can produce the required output without it having been anticipated; however, a classifier cannot since the correct output lies outside its set of classes.

\subsection{Describing Instead of Classifying}\label{sec:describing}
Previous approaches to automating Roman coin identification fall into two broad groups of classification-based approaches. The first predicts the catalogue type directly from the image~\citep{kim2016,guo2023}. The second recognizes individual elements of the design, such as the main figure on the reverse~\citep{anwar2021,cooper2020,reid2026}. Our approach instead generates a description of each side of the coin, comprising its legend and its imagery, from which the coin type can subsequently be identified. We argue that this formulation has clear advantages over both.

Generating a description has two advantages over classifying the type directly. First, it is less affected by the problem of class imbalance: some types are far more common than others and a type classifier sees few or no examples of the rare ones. Second, it makes the identification of the type of a coin more transparent: a classifier maps the image straight to a type, whereas our approach produces an intermediate textual description, from which the type is then determined by an expert or through a separate algorithm. Each component of that description can be inspected and, if wrong, corrected.

Compared with recognizing a single element, a full description is both more specific and more robust. A single element, such as the main figure depicted, rarely determines the type on its own, so a model that predicts only this element cannot by itself support automated cataloguing. Even where one element would suffice, the approach depends on that element being legible and fails when it is worn or damaged. A full description, by contrast, draws on all of the evidence on both sides of the coin, so damage to one element leaves the others usable and may still allow the type to be identified.

\subsection{Limitations}
The main limitation lies in the source data. The labels used for training and evaluation are derived from auction descriptions, which use inconsistent terminology and levels of detail, and often omit features visible on the coin. The obverse description labels additionally inherit the errors of the classifiers used to complete missing taxonomy components. Our results therefore measure agreement between the model's output and imperfect, sometimes incomplete, labels: the model is penalized for a correct output that does not match the label, and rewarded for reproducing the label's errors and omissions.

The evaluation is also narrow in two respects. The images come from auction records, which favour well-preserved coins photographed under consistent conditions. Excavation finds and museum specimens are often more worn and less uniformly photographed, so the performance reported here may not carry over to them. All results also come from a single base model and a single training run, so their sensitivity to the choice of backbone, model scale, and initialization is unknown.

\section{Conclusion and Future Work}\label{sec:conclusion}
This paper has formulated the task of describing Roman Imperial coins as autoregressive text generation, introduced metrics for evaluating the generated legends and descriptions, and shown that an open-weight VLM can be adapted to the task on modest hardware using training data derived automatically from auction records. The adapted models outperform both the base model and a proprietary general-purpose model on all four tasks, and their gains in legend transcription extend to inscriptions never seen in training. Together, these provide a foundation for generative approaches to Roman coin description, and we close by outlining the directions we consider most important.

The most direct improvement would be a dataset annotated for this task, in which experts record the legends and descriptions of both sides of each coin in a consistent format.

A second direction is to complete the identification process by adding a step that matches the generated descriptions to catalogue entries, for example through semantic similarity search, so that the pipeline outputs a catalogue type for each coin. This would then allow for a direct comparison between generation followed by search and direct classification, establishing which identifies coins more accurately.

A third direction concerns uncertainty. Future models could mark low-confidence or unreadable characters explicitly. They could also generate a reasoning trace indicating whether each character was read from the image or inferred from context, although such traces should be treated with caution, since a model's stated reasoning need not reflect the process that produced its prediction~\cite{turpin2023unfaithful}.

Finally, reinforcement learning could be used to further optimize the fine-tuned models, with the metrics defined in this work serving as the basis for reward functions. Our preliminary experiments with Group Relative Policy Optimization~\cite{shao2024grpo} were inconclusive, although they were limited by the group sizes our hardware allowed. A larger study could establish whether reinforcement learning improves on supervised fine-tuning alone.

\begin{acks}
We are grateful to ACSearch for supplying the auction data used in this work, and to Simon Wieland in particular.
\end{acks}

\bibliographystyle{ACM-Reference-Format}
\bibliography{references}

\appendix
\section{Task Prompts}\label{appendix:prompts}

\subsection{Structured Features Extraction Prompt}
\label{appendix:structured-features-extraction-prompt}
\begin{lstlisting}[style=promptstyle]
# ROLE
You are a precise Roman numismatic cataloguing assistant.

# INPUTS
You will receive:
1) An auction description of a Roman coin (may be EN/DE/FR/IT/ES, may contain abbreviations and line breaks)
2) An image showing BOTH sides of the coin (obverse + reverse)

# TASK
Extract, for EACH side:
- the LEGEND (inscription text) as written in the description
- a short DESCRIPTION as given in the auction description provided, translated to english

Use BOTH sources:
- Prefer the description for exact legend text and stated identifications.
- Use the image to disambiguate which side is obverse vs reverse and to confirm details when the text is vague.
- Use your knowledge about typical roman coin legends and the conventions when it comes to Roman coins.

# RULES
## Side assignment
- "Obverse" = side with the main portrait/head (usually the emperor/bust).
- "Reverse" = the other side (deity/scene/standing figure, etc.). Comes after the reverse legend.
If the description explicitly labels sides (e.g., "Obv.", "Rv."), follow that. Otherwise infer from the image.

## Legends (critical)
- Output legends in ALL CAPS.
- Copy the legend EXACTLY AS GIVEN IN THE DESCRIPTION, it must be a substring of the auction description provided.
- Do NOT reconstruct missing letters from the image.
- Do NOT expand abbreviations or correct spelling.
- If a legend is not present in the description provided or not clearly attributable to a side, output: "unknown".
- The legend must be included in the auction description provided

## Descriptions
- Output descriptions in ENGLISH.
- Base them on the provided description; translate if needed.
- Normalise formatting: remove line breaks, hyphenation from wraps, and odd separators. Produce a single clean sentence or short phrase.
- Keep it factual and concise (no speculation, no extra historical context).
- The obverse and reverse legend must always be different (unless they are both unknown)

## Uncertainty
- If any field is missing, ambiguous, or not confidently supported by the description, output "unknown" for that field (do not guess).

# OUTPUT FORMAT (JSON ONLY)
Return exactly one JSON object with these keys and string values:
{
"obverse_legend": "...",
"obverse_description": "...",
"reverse_legend": "...",
"reverse_description": "..."
}
No extra keys. No markdown. No commentary.
\end{lstlisting}

\subsection{Legend Extraction Prompt}
\label{appendix:legend-extraction-prompt}
\begin{lstlisting}[style=promptstyle]
# PERSONA
You are an OCR-style extraction module for ancient Roman coins.
You do not describe, analyse, classify, guess context, identify the emperor, translate, or explain. 
You only read and transcribe the legend text struck on the coin.

# TASK
1. Locate the coin in the image. A Roman coin is usually a circular or slightly irregular metal disc with a portrait, deity, figure or symbol in the centre.
2. Look at the outer ring of that disc. The legend is the sequence of large capital letters stamped near the rim of the coin, following the curve of the edge. It often wraps around the coin in a circle or arc.
3. Read those letters and transcribe them.

## WHAT TO EXTRACT
1. Extract only the legend text physically stamped on the coin surface.

# OUTPUT RULES
1. Output ONLY the legend text as a single line.
2. Use ONLY uppercase A-Z.
3. If the coin is worn or broken and a letter cannot be read with confidence, replace that letter with a space.
4. If you can only read a fragment, output that fragment.
5. If you cannot read any letters at all, output "unknown", nothing else.
6. You MUST follow these constraints:
   - All caps.
   - No quotes.
   - No reasoning or commentary.

# SHAPE AND LAYOUT GUIDANCE
Use these visual cues to locate the correct text:
1. Focus on the curved text that follows the outer edge of the coin. The text is usually arranged in an arc or full circle around the portrait or central figure.
2. The legend letters are often tall, bold, and slightly irregular, with serif-like strokes, and may be separated by dots or stars on the physical coin. Ignore those separators in the output.
\end{lstlisting}

\subsection{Obverse Description Extraction Prompt}
\label{appendix:obverse-description-extraction-prompt}
\begin{lstlisting}[style=promptstyle][t]
# PERSONA
You are a deterministic Roman obverse portrait extraction module.
Extract only what is visually present. Do not add commentary.

# TASK
From the obverse image, output one single-line structured description that matches this schema:
<emperor>; <headwear> <head/bust> facing <left/right>, <dress>[, <attribute1>, <attribute2>, ...]

# ALLOWED VALUES
Use only these label vocabularies:

1) HEADWEAR (exactly one)
- bare
- laureate
- radiate
- diademed
- helmeted
- veiled

2) HEAD OR BUST (exactly one)
- head
- bust

3) DIRECTION (exactly one)
- left
- right

4) DRESS (exactly one)
- draped
- cuirassed
- draped and cuirassed
- drapery on left shoulder
- consular robes
- aegis

5) ATTRIBUTES (zero or more, comma-separated, each at most once)
- holding spear
- holding spear and shield on shoulder
- holding sceptre
- holding eagle-topped sceptre
- holding globe
- holding Victory on globe

# OUTPUT RULES
- Single line only.
- No quotes, markdown, or extra text.
- Keep the literal word "facing" before direction.
- Use comma separators exactly as in:
  Emperor; headwear head_or_bust facing direction, wear, attr1, attr2
- If no attributes are visible, stop after wear:
  Emperor; headwear head_or_bust facing direction, wear
- If emperor is uncertain, use "unknown" as emperor token.
- If any non-emperor slot is uncertain, choose the closest valid label from the allowed list.

# EXAMPLES
Augustus; laureate head facing right, draped
Trajan; laureate bust facing right, draped and cuirassed, holding spear
unknown; diademed bust facing left, cuirassed, holding globe
\end{lstlisting}

\subsection{Reverse Description Extraction Prompt}
\label{appendix:reverse-description-extraction-prompt}
\begin{lstlisting}[style=promptstyle]
# PERSONA
You are a deterministic Roman coin reverse extraction module.
Extract only what is visually present.
Do NOT infer identity, symbolism, or historical meaning.
Do NOT describe the obverse.

# TASK
From the reverse image, produce a concise numismatic description using ONLY clearly visible elements.

# EXTRACTION STRUCTURE (in order)

1. MAIN TYPE (required if visible)
   - Deity or personification ONLY if visually certain.
   - Otherwise use neutral labels: standing male figure, seated female figure,
     emperor standing, captive, trophy, altar, temple, eagle, standard, etc.

2. POSTURE / ACTION (required if visible)
   - standing, seated, advancing, kneeling, holding, sacrificing, crowning, etc.

3. ATTRIBUTES / OBJECTS (optional, only if clearly visible)
   - holding spear, shield, patera, cornucopia, wreath, globe,
     with captive at feet, between standards, altar to left, etc.

4. ORIENTATION (required if visible)
   - right, left, facing, facing each other, etc.

# STYLE RULES
- Compact catalogue noun-phrase style.
- No full sentences.
- No legend transcription.
- No interpretation or historical context.
- Only visually certain elements.

# FORMAT RULES
- Single line only.
- No quotes.
- No trailing punctuation.

# EXAMPLES
Victory standing left, holding wreath
Standing male figure right, holding spear
Temple facade with four columns
\end{lstlisting}

\subsection{Reverse Description Entity Extraction Prompt}
\label{appendix:reverse-description-entity-extraction-prompt}
\begin{lstlisting}[style=promptstyle]
You extract structured information from Roman coin reverse descriptions.

Return ONLY:
- main_entity: short noun phrase for the primary figure/type (e.g., "Victory", "Hercules", "male figure"). Do not include any verbs. At most 3 words.
- main_objects: visible objects/attributes explicitly mentioned (e.g. "wreath", "palm", "scepter"). Ignore legends mentioned.
Rules:

- Be faithful to the text; do not add external historical knowledge.
- If uncertain, leave the list empty or the string blank.
- Keep questions concise and factual.
- Keep answers short and directly grounded in the description.
\end{lstlisting}
\end{document}